\documentclass[conference]{IEEEtran}
\IEEEoverridecommandlockouts
\usepackage{cite}
\usepackage{amsmath,amssymb,amsfonts}
\usepackage{algorithmic}
\usepackage{graphicx}
\usepackage{textcomp}
\usepackage{xcolor}
\usepackage{tikz,pgfplots}
\usepackage[utf8]{inputenc}
\usepackage{subcaption}
\usepackage{graphicx}
\usetikzlibrary{positioning}
\usepackage{siunitx}
\usepackage{textcomp}
\newcommand{\figurewidth}{3cm}
\newcommand{\figureheight}{1.8cm}
\usepackage{amsmath}
\usepackage{amsfonts}
\usepackage{amssymb}
\usepackage{booktabs}
\usepackage{subcaption}
\usepackage{hyperref}
\hypersetup{
    colorlinks=true,
    linkcolor=blue,
    filecolor=magenta,      
    urlcolor=cyan,
}
\usepackage{listings}
\usepackage{color}
\usepackage{wasysym}
\usepackage{algorithm}
\usepackage{algorithmic}
\usepackage{eucal}
\usepackage{bm}

\usepackage{enumitem}

\DeclareMathOperator*{\argmin}{arg\,min}
\definecolor{dkgreen}{rgb}{0,0.6,0}
\definecolor{gray}{rgb}{0.5,0.5,0.5}
\definecolor{mauve}{rgb}{0.58,0,0.82}

\definecolor{dkgreen}{rgb}{0,0.6,0}
\definecolor{gray}{rgb}{0.5,0.5,0.5}
\definecolor{mauve}{rgb}{0.58,0,0.82}

\renewcommand{\algorithmiccomment}[1]{\bgroup\hfill//~#1\egroup}

\newcommand{\model}{\textbf{AutoOD}~}
\newcommand{\modele}{\textbf{AutoOD}}

\newcommand{\nop}[1]{}
\usepackage{color}

\usepackage{multirow}

\AtBeginDocument{%
  \providecommand\BibTeX{{%
    \normalfont B\kern-0.5em{\scshape i\kern-0.25em b}\kern-0.8em\TeX}}}

\def\BibTeX{{\rm B\kern-.05em{\sc i\kern-.025em b}\kern-.08em
    T\kern-.1667em\lower.7ex\hbox{E}\kern-.125emX}}

\begin{document}
\title{AutoOD: Automated Outlier Detection via Curiosity-guided Search and Self-imitation Learning}

\title{AutoOD: Automated Outlier Detection via Curiosity-guided Search and Self-imitation Learning}


\author{\IEEEauthorblockN{Yuening Li$^{1}$, Zhengzhang Chen$^{2}$, Daochen Zha$^{1}$, Kaixiong Zhou$^{1}$, Haifeng Jin$^{1}$, Haifeng Chen$^{2}$, Xia Hu$^{1}$}
\IEEEauthorblockA{\textit{$^{1}$Department of Computer Science and Engineering, Texas A\&M University}, USA \\
\{yueningl,daochen.zha,zkxiong,jin,xiahu\}@tamu.edu\\
\textit{$^{2}$NEC Laboratories America}, USA\\
\{zchen,haifeng\}@nec-labs.com}
}
\maketitle



\begin{abstract}
Outlier detection is an important data mining task with numerous applications such as intrusion detection, credit card fraud detection, and video surveillance. However, given a specific complicated task with complicated data, the process of building an effective deep learning based system for outlier detection still highly relies on human expertise and laboring trials. Also, while Neural Architecture Search (NAS) has shown its promise in discovering effective deep architectures in various domains, such as image classification, object detection and semantic segmentation, contemporary NAS methods are not suitable for outlier detection due to the lack of intrinsic search space, unstable search process, and low sample efficiency. 
To bridge the gap, in this paper, we propose \modele, an automated outlier detection framework, which aims to search for an optimal neural network model within a predefined search space. Specifically, we firstly design a curiosity-guided search strategy to overcome the curse of local optimality. A controller, which acts as a search agent, is encouraged to take actions to maximize the information gain about the controller's internal belief. We further introduce an experience replay mechanism based on self-imitation learning to improve the sample efficiency. 
Experimental results on various real-world benchmark datasets demonstrate that the deep model identified by \model achieves the best performance, comparing with existing handcrafted models and traditional search methods.

\end{abstract}


\maketitle

\section{Introduction}



%

With the increasing amount of surveillance data collected from large-scale information systems such as the Web, 
social networks, and cyber-physical systems, it becomes more and more important for people to understand the underlying regularity of the vast amount of data, and to identify the unusual or abnormal instances~\cite{padmanabhan2013graph,Lin2018,li2019deep}. Centered around this goal, outlier detection plays a very important role in various real-world applications, such as fraud detection~\cite{chandola2009anomaly,cheng2016}, cyber security~\cite{ting2016,dong2017,wang2019attentional}, medical diagnosis~\cite{lin2005approximations}, and social network analysis~\cite{chen2012,li2019specae,huang2019graph,liu2019single}. 


Driven by the success of deep learning, there has been a surge of interests~\cite{guo2017deep,wang2019heterogeneous,zong2018deep,ruff2018deep} in adopting deep neural networks for outlier detection. Deep neural networks can learn to represent the data as a nested hierarchy of concepts to capture the complex structure in the data, and thus significantly surpass traditional outlier detection methods as the scale of data increases~\cite{hendrycks2018deep}. However, 
building a powerful deep neural network system for a real-world complex application usually still heavily relies on human expertise to fine-tune the hyperparameters and design the neural architectures.  
These efforts are usually time-consuming and the resulting solutions may still have sub-optimal performance.

Neural Architecture Search (NAS)~\cite{zoph2016neural,elsken2019neural} is one promising means for automating the design of neural networks, where reinforcement learning and evolution have been used to discover
optimal model architectures from data~\cite{cubuk2019autoaugment,pham2018efficient}. Designing an effective NAS algorithm requires two key components: the search space and the search strategy, which define what architectures can be represented in principles and how to explore the search space, respectively. The discovered neural architectures by NAS have been demonstrated to be on par or outperforms hand-crafted neural architectures. 

Although the recent years have witnessed significant progress of NAS techniques in some supervised learning tasks such as image classification and text classification~\cite{pham2018efficient,zoph2016neural}, the unsupervised setting and the naturally imbalanced data have introduced new challenges in designing an automated outlier detection framework. 
(1) \textbf{Lack of search space}. It is non-trivial to determine the search space for an outlier detection task. 
In particular, since there is no class label information in the training data of an outlier detection task, objective functions play an important role to differentiate between normal and anomalous behaviors. Thus, in contrast to the supervised learning tasks, we often need to find a suitable definition of the outlier and its corresponding objective function for a given real-world data. One typical way to define the outliers is to estimate the relative density of each sample, and declare instances that lie in a neighborhood with low density as anomalies~\cite{zong2018deep}. Yet these density-based techniques perform poorly if the data have regions of varying densities. Another way to define anomalies is through clustering. An instance will be classified as normal data if it is close to the existing clusters, while the anomalies are assumed to be far away from any existing clusters~\cite{guo2017deep}. However, these clustering-based techniques will be less effective if the anomalies form significant clusters among themselves~\cite{chandola2009anomaly}. 
The proper definition of outliers not only requires domain knowledge from researchers and experience from data scientists, but also needs thorough and detailed raw data analysis efforts. Thus, different from the search spaces defined by the existing NAS, the search space of automated outlier detection needs to cover not only the architecture configurations, but also the outlier definitions with corresponding objective functions. 
(2) \textbf{Unstable search process}. The search process may easily become unstable and fragile when outlier detection compounds with architecture search. On the one hand, the imbalanced data distributions make the search process easily fall into the local optima~\cite{swirszcz2016local}. On the other hand, the internal mechanisms of the traditional NAS may introduce bias in the search process. For instance, the weight sharing mechanism makes the architectures who have better initial performance with similar structures more likely to be sampled~\cite{chu2019fairnas,bender2020can}, which leads to misjudgments of the child model's performance. 
(3) \textbf{Low sample efficiency}. Existing NAS algorithms usually require training a large number of child models to achieve good performance, which is computationally expensive. While in real-world outlier detection tasks, outliers or abnormal samples are very rare. Thus, it requires the search strategy to exploit samples and historical search experiences more effectively.

To tackle the aforementioned challenges, in this paper, we propose \modele, an \underline{auto}mated \underline{o}utlier \underline{d}etection algorithm to find an optimal deep neural network model for a given dataset. In particular, we first design a comprehensive search space specifically tailored for outlier detection. It covers architecture settings, outlier definitions, and corresponding loss functions. Given the predefined search space, we further propose a curiosity-guided search strategy to overcome the curse of the local optimality. The search agent is encouraged to seek out regions in the search space that are relatively unexplored. The uncertainty about the dynamics of the search agent is interpreted as the information gain between the agent's new belief over the old one. Moreover, we introduce an experience replay mechanism based on self-imitation learning to enhance sample efficiency. It can benefit the search process through exploiting good experience in the historical episodes. To evaluate the performance of \modele, we perform an extensive set of experiments on eight benchmark datasets. When tested on the two important outlier detection tasks---instance-level outlier sample detection and pixel-level defect region segmentation---our algorithm demonstrated the superior performance, comparing with existing handcrafted models and traditional search methods. The experimental results also show that AutoOD has the potential to be applied in more complicated real-world applications.

The contributions of this paper are summarized as follows:
\begin{itemize}[wide=0pt, leftmargin=\dimexpr\labelwidth + 2\labelsep\relax]
   \item We identify a novel and challenging problem (\textit{i.e.}, automated outlier detection) and propose a generic framework \modele. To the best of our knowledge, AutoOD describes the first attempt to incorporate AutoML with an outlier detection task, and one of the first to extend AutoML concepts into applications from data mining fields.
    \item We carefully design a \textbf{search space} specifically tailored to
    the automated outlier detection problem, 
    covering architecture settings, outlier definitions, and the corresponding objective functions. 
    \item We propose a curiosity-guided search strategy to overcome the curse of 
    local optimality and \textbf{stabilize search process}.
    \item We introduce an experience replay mechanism based on the self-imitation learning to improve the sample efficiency.
    \item We conduct extensive experiments on eight benchmark datasets to demonstrate the effectiveness of \modele, and provide insights on how to incorporate \model to the real-world scenarios.
    
  
\end{itemize}

\begin{figure*}[t]
\centering
  \includegraphics[width=1.05\textwidth]{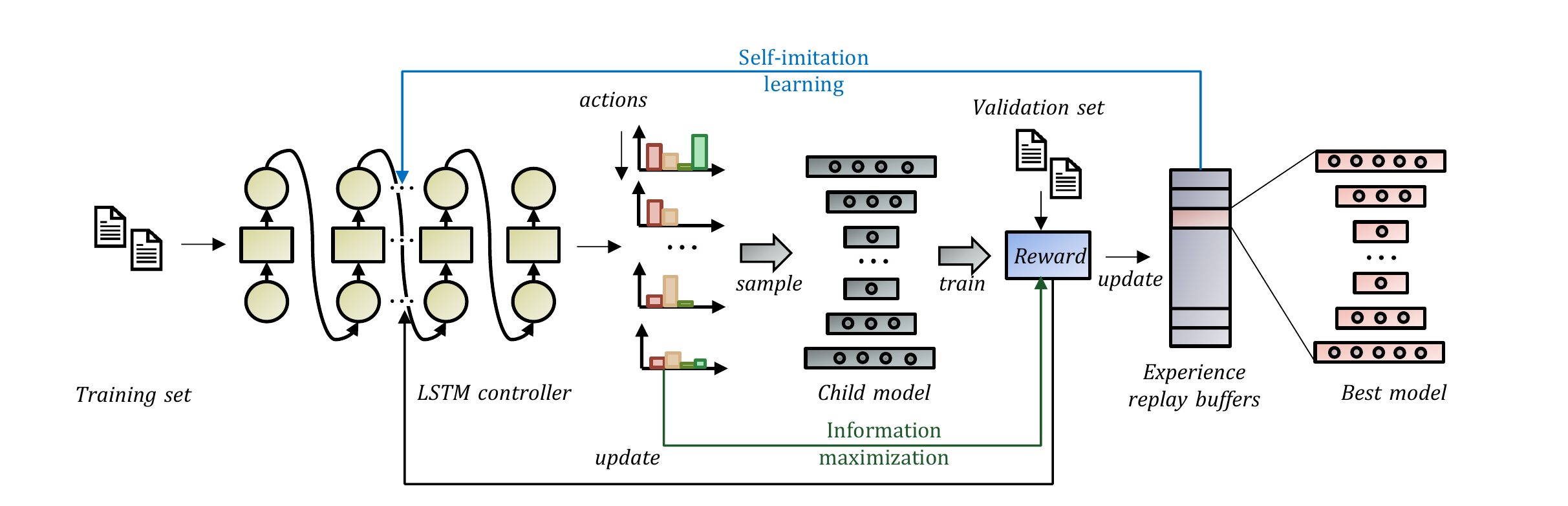}
  \caption{An overview of \modele. With the pre-defined search space and the given dataset, we use an LSTM based controller to generate actions $a$. Child models are sampled from actions $a$ and evaluated with the reward $r$. Once the search process of one iteration is done, the controller samples $M$ child models as candidate architectures and then
picks the top $K$ from them. The top $K$ architectures’ controller outputs will fed as the input of the next iteration’s controller.
  Parameters $\theta$ of the controller are updated with the reward $r$. $r$ is also shaped by information maximization about the controller's internal belief of the model, which is designed to guide the search process (shown as the green line). Good past experiences evaluated by the reward function are stored in replay buffers for future self-imitations (shown as the blue line).} 
  \label{fig:pipeline}
\end{figure*}

\section{Preliminaries and Problem Formulation}
\label{sec:pre}


In this section, we present the preliminaries and problem definition of our work.


\subsection{Deep AutoEncoder Based Outlier Detection}

Classical outlier detection methods, such as Local Outlier Factor~\cite{breunig2000lof} and One-Class SVMs~\cite{chen2001one}, suffer from bad computational scalability and the curse of dimensionality in high-dimensional and data-rich scenarios~\cite{ruff2018deep}. 
To tackle these problems, deep structured models have been proposed to process the features in a more efficient way. Among recent deep structured studies, Deep AutoEncoder is one of the most promising approaches for outlier detection. 
The AutoEncoder learns a representation by minimizing the reconstruction error from normal samples~\cite{zhou2017anomaly}. Therefore, it can be used to extract the common factors of variation from normal samples and reconstruct them easily, and vice versa. Besides directly employing the reconstruction error as the denoter, recent studies~\cite{guo2017deep,zong2018deep,ruff2018deep} demonstrate the effectiveness of collaborating Deep AutoEncoders with classical outlier detection techniques, by introducing regularizers through plugging learned representations into classical outlier definition hypotheses. Specifically, there are three typical outlier assumptions: density, cluster, and centroid. The density based approaches~\cite{zong2018deep} estimate the relative density of each sample, and declare instances that lie in a neighborhood with low density as anomalies. Under the clustering based assumption, normal instances belong to an existing cluster in the dataset, while anomalies are not contained in any existing cluster~\cite{guo2017deep}. The centroid based approaches~\cite{ruff2018deep} rely on the assumption that normal data instances lie close to their closest cluster centroid, while anomalies are far away from them. In this work, we illustrate the proposed \model by utilizing Deep AutoEncoder with a variety of regularizers as the basic outlier detection algorithm. The framework of \model could be easily extended to other deep-structured outlier detection approaches.

\subsection{Problem Statement}
Different from the traditional Neural Architecture Search, which focuses on optimizing neural network architectures for supervised learning tasks, automated outlier detection has the following two unique characteristics. First, the neural architecture in the Autoencoder needs to be adaptive in the given dataset to achieve competitive performance. The hyperparameter configurations of neural architecture include the number of layers, the size of convolutional kernels and filters, \textit{etc.}; Second, the outlier detection requires the designs of definition-hypothesis and corresponding objective function. Formally, we define the outlier detection model and the unified optimization problem of automated outlier detection as follows.

\noindent \textbf{Outlier Detection Model}: The model of outlier detection consists of three key components: the neural network architecture $A$ of AutoEncoder, the definition-hypothesis $H$ of outlier assumption, and the loss function $L$. We represent the model as a triple $(A, H, L)$.

\nop{\noindent \textbf{Automated Outlier Detection}: Let the triple $(\mathcal{A}, \mathcal{H}, \mathcal{L})$ denote the search space of outlier detection models, where $\mathcal{A}$ denotes the architecture subspace, $\mathcal{H}$ denotes the definition-hypothesis subspace, and $\mathcal{L}$ denotes the loss functions subspace. Given training set $\mathcal{D}_{\textrm{train}}$ and validation set
$\mathcal{D}_{\textrm{valid}}$, the \textbf{Combined neural Architecture SearcH and Definition-hypothesis Optimization(CASHDO)} aims to find the optimal model $(A^\star, H^\star, L^\star)$ to minimize the objective function $\mathcal{J}$ as follows:
\begin{equation}
     (A^\star, H^\star, L^\star) = \argmin_{A \in \mathcal{A}, H \in \mathcal{H}, L \in \mathcal{L}} \!\!\!\!\!\! \mathcal{J} (A(\omega), H, L, \mathcal{D}_{\textrm{train}}, \mathcal{D}_{\textrm{valid}} ).
\end{equation}}

\noindent \textbf{Automated Outlier Detection}: Let the triple $(\mathcal{A}, \mathcal{H}, \mathcal{L})$ denote the search space of outlier detection models, where $\mathcal{A}$ denotes the architecture subspace, $\mathcal{H}$ denotes the definition-hypothesis subspace, and $\mathcal{L}$ denotes the loss functions subspace. Given training set $\mathcal{D}_{\textrm{train}}$ and validation set
$\mathcal{D}_{\textrm{valid}}$, we aim to find the optimal model $(A^\star, H^\star, L^\star)$ to minimize the objective function $\mathcal{J}$ as follows:
\begin{equation}
     (A^\star, H^\star, L^\star) = \argmin_{A \in \mathcal{A}, H \in \mathcal{H}, L \in \mathcal{L}} \!\!\!\!\!\! \mathcal{J} (A(\omega), H, L, \mathcal{D}_{\textrm{train}}, \mathcal{D}_{\textrm{valid}} ),
\end{equation}
where $\omega$ denotes the weights 
well trained on architecture $A$. $\mathcal{J}$ denotes the loss on $\mathcal{D}_{\textrm{valid}}$ using the model trained on the $\mathcal{D}_{\textrm{train}}$ with definition-hypothesis $H$ and loss function $L$.

\section{Proposed Method}
\label{sec:method}
In this section, we propose an automated outlier detection framework to find the optimal neural network model for a given dataset. A general search space is designed to include the neural architecture hyperparameters, definition-hypothesis, and objective functions. To overcome the curse of local optimality under certain unstable search circumstances, we propose a curiosity-guided search strategy to improve search effectiveness. Moreover, we introduce an experience replay mechanism based on self-imitation learning to better exploit the past good experience and enhance the sample efficiency. An overview of \model is given in Fig. \ref{fig:pipeline}.

\nop{In this section, we propose a tailored automated framework to facilitate the design of outlier detection model. A general search space is designed to include the neural architecture hyperparameters, definition-hypothesis, and objective function. To overcome the curse of local optimality during the unstable search progress, we propose a curiosity-guided search strategy to improve the search effectiveness. Moreover, we introduce a self-imitation learning mechanism based on experience replay to better exploit past good experiences and enhance the sample efficiency. } 

\nop{In this section, we introduce the core idea of AutoOD, for automatically designing optimal model for outlier detection. AutoOD is built on two components: search space and search strategy. We further refine and break the search strategy into curiosity-guided search and self-imitation based learning processes. We describe the details of each components below.
Given the combined optimization problem, we propose a tailored automated framework to facilitate the design of outlier detection model. A general search space is designed to include the neural architecture hyperparameters, definition-hypothesis and objective function. To alleviate the curse of local optima during the unstable search progress, we propose a curiosity-guided search strategy to improve the search effectiveness. Moreover, we introduce a self-imitation learning mechanism based on experience replay to enhance the sample efficiency.  More details of the search space design and search strategies are introduced in the following sections.}

\subsection{Search Space Design}
 
 
Because there is a lack of intrinsic search space for outlier detection tasks, here we design the search space for the Deep AutoEncoder based algorithms, which is composed of global settings for the whole model, and local settings in each layer independently. Formally, we have:  

 \begin{equation}
  \begin{array}{l}
    A = \{f^{1}(\cdot),...,f^{N}(\cdot),g^{1}(\cdot),...,g^{N}(\cdot)\},\\
    f^{i}(x;\mathcal{\omega}_{i}) = \textrm{ACT}(\textrm{NORMA}(\textrm{POOL}(\textrm{CONV}(x))),\\
   g^{i}(x;\mathcal{\omega}_{i}) = \textrm{ACT}(\textrm{NORMA}(\textrm{UPPOOL}(\textrm{DECONV}(f(x)))),\\
    \textrm{score} = \textrm{DIST}(g(f(x;\mathcal{\omega})),x)+\textrm{DEFINEREG}(f(x;\mathcal{\omega})),
    \end{array}
\label{eq:operations}
\end{equation}
where $x$ denotes the set of instances as input data, and $\omega$ denotes the trainable weight matrix. The architecture space $A$ contains $N$ encoder-decoder layers. $f(\cdot)$ and $g(\cdot)$ denote encoder and decoder functions, respectively. 
$\textrm{ACT}(\cdot)$ is the activation function set. $\textrm{NORMA}$ denotes the normalization functions. $\textrm{POOL}(\cdot)$ and $\textrm{UPPOOL}(\cdot)$ are pooling methods. $\textrm{CONV}(\cdot)$ and $\textrm{DECONV}(\cdot)$ are convolution functions. As we discussed in the Section II (A), the encoder-decoder based outlier score $\textit{score}$ contains two terms: a reconstruction distance and an outlier regularizer. $\textrm{DIST}(\cdot)$ is the metric to measure the distance between the original inputs and the reconstruction results. $\textrm{DEFINEREG}(\cdot)$ acts as an regularizer to introduce the definition-hypothesis from $H$.  We revisit and extract the outlier detection hypotheses and their mathematical formulas from state-of-the-art approaches as shown in the Table~\ref{tab:concepts}. We decompose the search space defined in Eq.~\ref{eq:operations} into the following $8$ classes of actions:

\begin{table}[t]
\tiny
\begin{tabular}{@{}c|l@{}}
\toprule
Definitions $H$   & Regularizer Equations              \\ \midrule
Density~\cite{zong2018deep}          &  $-\log \left(\sum_{k=1}^{K} \hat{\phi}_{k} \frac{\exp \left(-\frac{1}{2}\left(f\left(x_{i} ; \mathcal{\omega}\right)-\hat{\mu}_{k}\right)^{T} \hat{\mathbf{\Sigma}}_{k}^{-1}\left(f\left(x_{i} ; \mathcal{\omega}\right)-\hat{\mu}_{k}\right)\right)}{\sqrt{\left|2 \pi \hat{\mathbf{\Sigma}}_{k}\right|}}\right)$\\ 
Cluster~\cite{guo2017deep}  &   $\sum_{i} \sum_{j} p_{i j} \log p_{i j}\left({\frac{\left(1+\left\|f\left(x_{i}; \mathcal{\omega}\right)-\mu_{j}\right\|^{2}\right)^{-1}}{\sum_{j}\left(1+\left\|f\left(x_{i} ; \mathcal{\omega}\right)-\mu_{j}\right\|^{2}\right)^{-1}}}\right)^{-1}$                  \\
Centroid~\cite{ruff2018deep}   &    $R^{2}+ \sum_{i=1}^{n} \max \left\{0,\left\|f\left(x_{i} ; \mathcal{\omega}\right)-c\right\|^{2}-R^{2}\right\}$                    \\
Reconstruction~\cite{zhou2017anomaly}   &     $\frac{1}{n} \sum_{i=1}^{n}\left\|g\left(f\left(x_{i} ; \mathcal{\omega}\right)\right)-x_{i}\right\|_{2}^{2}$               \\ \bottomrule
\end{tabular}
\caption{The set of four representative outlier detection hypotheses, where $f(\cdot)$ and $g(\cdot)$ denote encoder and decoder functions, respectively.} 
\label{tab:concepts}
\end{table}

\begin{itemize}[wide=0pt, leftmargin=\dimexpr\labelwidth + 2\labelsep\relax]
\item[] \textbf{Global Settings:}
\item {\verb|Definition-hypothesis|} determines the way to define the ``outliers'', which acts as a regularization term in the objective functions. We consider density-based, cluster-based, centroid-based, and reconstruction-based assumptions, as shown in Table~\ref{tab:concepts}.


\item {\verb|Distance measurement|} stands for the matrix measuring the distance for the reconstruction purpose, including 
$l_1$, $l_2$, $l_{2,1}$ norms, and the structural similarity (SSIM).

\item[] \textbf{Local Settings in Each Layer:}

\item {\verb|Output channel|} is the number of channels produced by the convolution operations in each layer, \textit{i.e.}, {3, 8, 16, 32, 64, 128, and 256}.

\item {\verb|Convolution kernel|} denotes the size of the kernel produced by the convolution operations in each layer, \textit{i.e.}, $1 \times 1, 3 \times 3, 5 \times 5,$ and $7 \times 7$.

\item {\verb|Pooling type|} denotes the type of pooling in each layer, including the max pooling and the average pooling.

\item {\verb|Pooling kernel|} denotes the kernel size 
of pooling operations in each layer, \textit{i.e.}, $1 \times 1, 3 \times 3, 5 \times 5,$ and $7 \times 7$.s

\item {\verb|Normalization type|} denotes the normalization type in each layer, including three options: batch normalization, instance normalization, and no normalization.

\item {\verb|Activation function|} is a set of activation functions in each layer, including Sigmoid, Tanh, ReLU, Linear, Softplus, LeakyReLU, ReLU6, and ELU.

\end{itemize}

\begin{figure}[t]
  \includegraphics[width=0.5\textwidth]{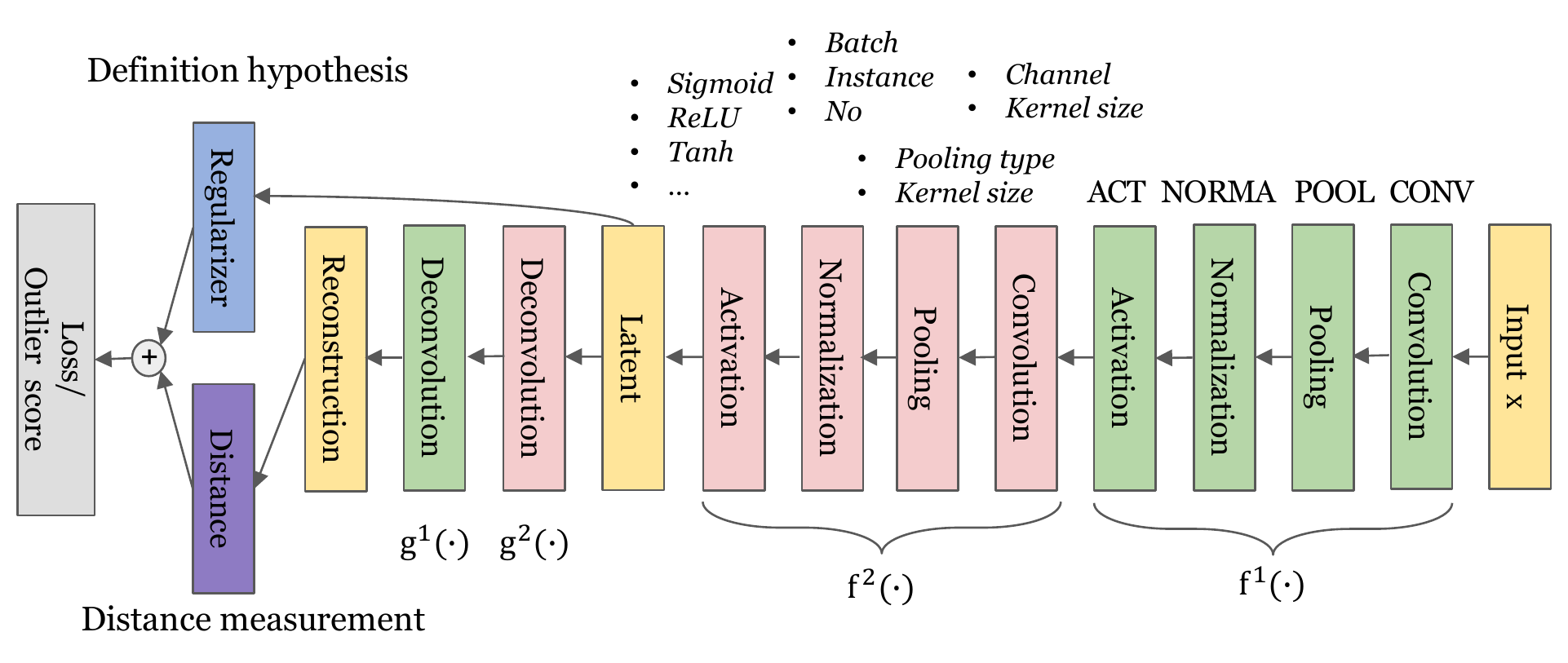}
  \caption{An example of the search space in AutoOD with two layers, which is composed of global settings for the whole model (blue and purple parts), and local settings in each layer (red and green parts), respectively. All building blocks are wired together to form a direct acyclic graph.} 
  \label{fig:searchspace}
\end{figure}

Thus, we use a $(6N+2)$ element tuple to represent the model, where $N$ is the number of layers in the encoder-decoder-wise structure.  Our search space includes an exponential number of settings. Specifically, if the encoder-decoder cell has $N$ layers and we allow action classes as above, it provides $4 \times 4 \times (7 \times 4 \times 2 \times 4 \times 3 \times 8)^N$ possible settings. Suppose we have a $N=6$, the number of points in our search space is $3.9e+23$, which requires an efficient search strategy to find an optimal model out of the large search space. Fig.~\ref{fig:searchspace} illustrates an example of the proposed search space in AutoOD.

\subsection{Curiosity-guided Search}
We now describe how to search the optimal model within the given search space. Inspired by the recent NAS work, the search strategy is considered as a meta-learning process. A controller is introduced to explore a given search space by training a child model to get an evaluation for guiding exploration~\cite{pham2018efficient}. The controller is implemented as a recurrent neural network. We use the controller to generate a sequence of actions for the child model. The whole process can be treated as a reinforcement learning problem with an action $a_{1:T}$, and a reward function $r$. 
To find the optimal model, we ask our controller to maximize its expected reward $r$, which is the expected performance in the validation set of the child models. 

There are two sets of learnable parameters: one of them is the shared parameters of the child models, denoted by $\omega$, and the other one is from the LSTM controller, denoted by $\theta$. $\omega$ is optimized using stochastic gradient descent (SGD) with the gradient $\nabla_{\omega}$ as:
\begin{equation}
    \label{eq:enasomega}
    \nabla_{\omega}\mathbb{E}_{m \sim \pi(m;\theta)}[L(m;\omega)] \approx \nabla_{\omega}L(m,\omega),
\end{equation}
where child model $m$ is sampled from the controller’s actions $\pi(m;\theta)$, $L(m,\omega)$ is the loss function composed from the search space above, computed on a minibatch of training data. The gradient is estimated using the Monte Carlo method.

Since the reward signal $r$ is non-differentiable, to maximize the expected reward $r$, we fix $\omega$ and apply the REINFORCE rule~\cite{williams1992simple} to update the controller's parameters $\theta$ as:  
\begin{equation}
    \label{eq:enastheta}
    \nabla_{\theta}\mathbb{E}_{P(a_{1:t};\theta)}[r\nabla_{\theta} \log P(a_t|a_{1:t-1};\theta)],
\end{equation}
where $r$ is computed as the performance on the validation set, rather than on the label-free training set. We define the reward $r$ as the detection accuracy of the sampled child model. We also adopt different evaluation metrics, including AUROC, AUPR, and RPRO in the experiment section. An empirical approximation of the Eq.~\ref{eq:enastheta} is:
\begin{equation}
    \label{eq:enasthetaappr}
    L = \frac{1}{n}\sum_{k=1}^{n}\sum_{t=1}^{T} (r_k-b)\nabla_{\theta}\log P(a_t|a_{1:t
    -1};\theta), 
\end{equation}
where $n$ is the number of different child models that the controller samples in one batch and $T$ is the number of tokens. $b$ acts as a baseline function  to reduce the estimate variance. 

\subsubsection{Curiosity-driven Exploration}
Despite being widely utilized due to search efficiency, weight sharing approaches are roughly built on empirical experiments instead of solid theoretical ground~\cite{chu2019fairnas}. The unfair bias will make the controller misjudge the child-model performance: those who have better initial performance with similar child models are more likely to be sampled. In the meanwhile, due to the imbalanced label distribution in outlier detection tasks, it is easy to make the controller fall into local optima. 


To address these problems, \model builds on the theory of curiosity-driven exploration~\cite{sun2011planning}, aiming to encourage the controller to seek out regions in the searching spaces that are relatively unexplored. It brings us a typical exploration-exploitation dilemma to guide the controller.

Bayesian reinforcement learning~\cite{fortunato2017bayesian,ghavamzadeh2015bayesian} offers us a formal guarantees as coherent probabilistic model for reinforcement learning. It provides a principled framework to express the classic exploration-exploitation dilemma, by keeping an explicit representation of uncertainty, and selecting actions that are optimal with respect to a version of the problem that incorporates this uncertainty~\cite{ghavamzadeh2015bayesian}. Here, instead of a vanilla RNN, we use a Bayesian LSTM as the structure of the controller to guide the search. The controller’s understanding of the search space is represented dynamically over the uncertainty of  the parameters of the controller. Assuming a prior $p(\theta)$, it maintains a distribution prior over the controller's parameters through a distribution over $\theta$. The controller models the actions via $p(a_t|a_{1:t};\theta)$, paratetrized by $\theta$. According to curiosity-driven exploration~\cite{sun2011planning}, the uncertainty about the controller's dynamics
can be formalized as maximizing the information:
\begin{equation}
\label{informationgain}
I(a_{t};\theta|a_{1:{t-1}})=\mathbb{E}_{a_{t}\sim P(\cdot|a_{1:{t-1}})}\big[D_{\mathrm{KL}}[p(\theta|a_{1:{t-1}})\, || \,p(\theta)]\big],
\end{equation}
where the KL divergence can be interpreted as \textit{information gain}, which denotes the mutual information between the controller's new belief over the model to the old one.

Thus, the information gain of the posterior dynamics distribution of the controller can be approximated as an \textit{intrinsic reward}, which captures the controller’s surprise in the form of a reward function. We can also use the REINFORCE rule to approximate planning for maximal mutual information by adding the intrinsic reward along with the external reward (accuracy on the validation set) as a new reward function. It can also be interpreted as a trade-off between exploitation and exploration as:
\begin{equation}
\label{eq:reward}
r_{\textrm{new}}(a_{t})= r(a_{t})+\eta D_{\mathrm{KL}}[p(\theta|a_{1:{t-1}})\, || \,p(\theta)],
\end{equation}
where $\eta \in \mathbb{R}_{+}$ is a hyperparameter controlling the urge to explore. However, it is generally intractable to calculate the posterior $p(\theta|a_{1:{t-1}})$ in Eq.~\ref{eq:reward}.

\subsubsection{Variational Bayes-by-Backprop}
In this subsection, we propose a tractable solution to maximize the information gain objective presented in the previous subsection. To learn a probability distribution over network parameters $\theta$,  we propose a practical solution through a back-propagation compatible algorithm, \textit{bayes-by-backprop}~\cite{blundell2015weight,fortunato2017bayesian}.

In Bayesian models, latent variables are drawn from a prior density $p(\theta)$. During inference, the posterior distribution $p(\theta|x)$ is computed given a new action through Bayes' rule:
\begin{equation}
\label{bayesrule} 
p(a_{t}|a_{1:t-1})=\frac{p(\theta)p(a_{t}|a_{1:t-1};\theta)}{p(a_{t}|a_{1:t-1})}.
\end{equation}
The denominator can be computed through the integral:
\begin{equation}
\label{integralbayes} 
p(a_{t}|a_{1:t-1})=\int_{\Theta}p(a_{t}|a_{1:t-1};\theta)p(\theta)\mathrm{d}\theta.
\end{equation}
As controllers are highly expressive parametrized LSTM networks, which are usually intractable as high-dimensionality. Instead of calculating the posterior $p(\theta|\mathcal{D_{\textrm{train}}})$ for a training dataset $\mathcal{D_{\textrm{train}}}$. We approximate the posterior through an alternative probability densities over the latent variables $\theta$ as $q(\theta)$, by minimizing the Kullback-Leibler(KL) divergence $D_\mathrm{KL}[q(\theta)\,||\,p(\theta)]$. We use $\mathcal{D}$ instead of $\mathcal{D_{\textrm{train}}}$ in the following parts of this subsection for brevity.
\begin{equation}
\label{qgaussiandistribution}
    q(\theta)=\prod_{i=1}^{|\Phi|}\mathcal{N}(\theta_i|\mu_i;\sigma_{i}^2).
\end{equation}
\nop{
\begin{equation}
\label{qgaussiandistribution}
    q(\theta;\phi)=\prod_{i=1}^{|\Phi|}\mathcal{N}(\theta_i|\mu_i;\sigma_{i}^2), 
    \phi=\{\mu,\sigma\}.
\end{equation}}
$q(\theta)$ is given by a Gaussian distribution, with $\mu$ as the Gaussian's mean vector and $\sigma$ as the covariance matrix.


Once minimized the KL divergence, $q(\cdot)$ would be the closest approximation to the true posterior. Let $\log p(\mathcal{D}|\theta)$ be the log-likelihood of the model. Then, the network can be trained by minimizing the variational free energy as the expected lower bound:
\begin{equation}
\label{variationallowerbound}
L[q(\theta),\mathcal{D}] = - \mathbb{E}_{\theta \sim q(\cdot )}\big[\log p(\mathcal{D}|\theta)\big] + D_{\mathrm{KL}}\big[q(\theta)\, || \,p(\theta)\big],
\end{equation}
which can be approximated using $N$ Monte Carlo samples from the variational posterior with $N$ samples drawn according to $\theta \sim q(\cdot)$: 
\begin{equation}
\label{variationallowerboundestimate}
L[q(\theta),\mathcal{D}] \approx \sum_{i=1}^{N} -\log  p(\mathcal{D}|\theta^{(i)}) + \log q(\theta^{(i)})-\log p(\theta^{(i)}).
\end{equation}
\subsubsection{Posterior Sharpening}
We discuss how to derive a distribution $q(\theta|\mathcal{D})$ to improve the gradient estimates of the intractable likelihood function $p(\mathcal{D})$, which is related to Variational AutoEncoders (VAEs)~\cite{kingma2013auto}. Inspired from strong empirical evidence and extensive work on VAEs, the ``sharpened'' posterior yields more stable optimization. We now use posterior sharpening strategy to benefit our search process. 

The challenging part of modelling the variational posterior $q(\theta|\mathcal{D})$ is the high dimensional $\theta \in \mathbb{R}^d$, which makes the modelling unfeasible. Given the first term of the loss $-\log p(\mathcal{D}|\theta)$ is differentiable with respect to $\theta$, we propose to parameterize $q$ as a linear combination of $\theta$ and $-\log p(\mathcal{D}|\theta)$. Thus, we can define the hierarchical posterior of the form in Eq.~\ref{qgaussiandistribution}:
\begin{equation}
\label{hierachicalposterior}
    q(\theta|\mathcal{D})=\int q(\theta|\phi,\mathcal{D})q(\phi)\mathrm{d}\phi,
\end{equation}
\begin{equation}
\label{hierachicalposteriorreparameter}
    q(\theta|\phi, \mathcal{D})= \mathcal{N}(\theta|\phi-\eta*-\nabla_{\phi}\log   p(\mathcal{D}|\phi), \sigma^{2}I)
\end{equation}
with $\mu, \sigma \in \mathbb{R}^{d}$, and $q(\phi)=\mathcal{N}(\phi|\mu,\sigma)$ as the same setting in the standard variational inference method. $\eta \in \mathbb{R}^{d}$ can be treated as a per-parameter learning rate.

In the training phrase, we have $\theta \sim q(\theta|\mathcal{D})$ via ancestral sampling to optimise the loss as:
\begin{equation}
\label{eq:traininglossbrnn}
    L_{\text{explore}}=L(\mu,\sigma,\eta) = \mathbb{E}_{\mathcal{D}}[\mathbb{E}_{q(\phi)q(\theta|\phi,\mathcal{D})}[L(\mathcal{D},\theta,\phi|\mu,\sigma,\eta)]]
\end{equation}
with $L(\mathcal{D},\theta,\phi|\mu,\sigma,\eta)$ given by:
\begin{equation}
\begin{aligned}
\label{eq:exploreloss2}
    L(\mathcal{D},\theta,\phi|\mu,\sigma,\eta) &= -\log p(\mathcal{D}|\theta) + \mathrm{KL}[q(\theta|\phi,\mathcal{D})\, || \,p(\theta|\phi)] \\
    & \quad+\frac{1}{C}\mathrm{KL}[q(\phi)\, || \,p(\phi)],
\end{aligned}
\end{equation}
where the constant $C$ is the number of  truncated sequences.

Thus, we turn to deriving the training loss function for posterior sharpening. With the discussion above, we assume a hierarchical prior for the parameters such that $p(\mathcal{D})=\int p(\mathcal{D}|\theta)p(\theta|\phi)p(\phi)\mathrm{d}\theta\mathrm{d}\phi$. Then, the expected lower bound on $p(\mathcal{D})$ is defined as follows:
\begin{equation}
\label{elbo}
    \begin{aligned}
     & \quad \log p(\mathcal{D}) \\
   &= \log \left(\int p(\mathcal{D}|\theta)p(\theta|\phi)p(\phi)\mathrm{d}\theta\mathrm{d}\phi \right) \\
    &\geqslant \mathbb{E}_{q(\phi,\theta|\mathcal{D})}\left[ \log \frac{p(\mathcal{D}|\theta)p(\theta|\phi)p(\phi)}{q(\phi,\theta|\mathcal{D})}\right] \\
    &= \mathbb{E}_{q(\theta|\phi,\mathcal{D})q(\phi)}\left[\log\frac{p(\mathcal{D}|\theta)p(\theta|\phi)p(\phi)}{q(\theta|\phi,\mathcal{D})q(\phi)} \right]\\
    &= \mathbb{E}_{q(\phi)} \left[ \mathbb{E}_{q(\theta|\phi,\mathcal{D})}\left[\log p(\mathcal{D}|\theta) + \log \frac{p(\theta|\phi)}{q(\theta|\phi,\mathcal{D})} \right] +\log \frac{p(\phi)}{q(\phi)} \right]\\
    &= \mathbb{E}_{q(\phi)} \Big[ \mathbb{E}_{q(\theta|\phi,\mathcal{D})} \big[\log p(\mathcal{D}|\theta)-\mathrm{KL}[q(\theta|\phi,\mathcal{D})\, || \,p(\theta|\phi)] \big] \\
    &\quad -\mathrm{KL}[q(\phi)\, || \,p(\phi)] \Big].
    \end{aligned}
\end{equation}         

\begin{algorithm}[t]
\small
\caption{Automated Outlier Detection}\label{algo:optimization}
\begin{algorithmic}[1]
   \STATE {\bfseries Input:} Input datasets $\mathcal{D}_{\textrm{train}},\mathcal{D}_{\textrm{valid}}$, and search space $\mathcal{S}$.
    \STATE {\bfseries Output:} Optimal model with the best performance.
    \STATE Initialize parameter $\theta, \omega$;
    \STATE Initialize replay buffer $\mathcal{B} \leftarrow \emptyset$;
    \FOR {each iteration}
    \STATE \textit{Perform curiosity-guided search via a LSTM controller}
    \FOR {each step $t$}
    \STATE Sample an action $a_t \sim \pi(a_{1:t-1};\theta)$;
    \STATE $\omega \leftarrow \omega-\eta\nabla_{\omega}\mathbb{E}_{a_t \sim \pi(a_{1:t-1};\theta)}[L(a_{1:t-1};\omega)]$; \COMMENT{Eq.~\ref{eq:enasomega}}

    \STATE $\theta \leftarrow \theta  - \eta L_{\textrm{explore}}(\mathcal{D}_{\textrm{train}},\theta)$ ; \COMMENT{Eq.~\ref{eq:traininglossbrnn}}
        \STATE  $r_{\textrm{new}}(a_{t}) \leftarrow  r(a_{t})+\eta D_{\mathrm{KL}}[p(\theta|a_{1:{t-1}})\, || \,p(\theta)]$;   \COMMENT{Eq.~\ref{eq:reward}}
    \STATE Update controller via the new reward  $r_{\textrm{new}}(a_{t})$; \COMMENT{Eq.~\ref{eq:enastheta}}
    \IF {the performance of $a_t$ on
    $\mathcal{D}_{\textrm{val}}$ outperforms the actions stored in $\mathcal{B}$} 
    \STATE {$\mathcal{B} \leftarrow \{\bm{a}, r\} \cup \mathcal{B} $}; Update replay buffer;
    \ENDIF
    \ENDFOR
    \STATE {\textit{Perform self-imitation learning}}
    \FOR {each step t}
        \STATE Sample a mini-batch $\{\bm{a}, r\}$ from $\mathcal{B}$;
        \STATE $\omega \leftarrow \omega-\eta\nabla_{\omega}\mathbb{E}_{a_t \sim \pi(a_{1:t-1};\theta)}[L(a_{1:t-1};\omega)]$; \COMMENT{Eq.~\ref{eq:enasomega}}
        \STATE $\theta \leftarrow \theta  - \eta L_{\textrm{replay}}(\mathcal{D}_{\textrm{valid}},\theta)$; \COMMENT{Eq.~\ref{eq:simappr}}
    \ENDFOR
    \ENDFOR
\end{algorithmic}
\end{algorithm}

\subsection{Experience Replay via Self-Imitation Learning}
The goal of this subsection is to exploit the past good experiences for the controller to benefit the search process by enhancing the sample efficiency, especially considering there are only a limited number of negative samples in outlier detection tasks. In this paper, we propose to store rewards from historical episodes into experience replay buffers ~\cite{oh2018self}: $ \mathcal{B} = {(a_{1:t}, r_{\bm{a}})}$, where $(a_{1:t}$ and $r_{\bm{a}})$ are the actions and the corresponding reward. To exploit good past experiences, we update the experience replay buffer for child models with better rewards, and amplify the contribution from them to the gradient of $\theta$. More specifically, we sample child models from the replay buffer using the clipped advantage $(r-b)_{+}$, where the rewards $r$ in the past experiences outperform the current baseline $b$. Comparing with the Eq.~\ref{eq:enasthetaappr}, the objective to update the controller's parameter $\theta$ through the replay buffer is:

\begin{equation}
    \label{eq:sim}
    \nabla_{\theta}\mathbb{E}_{{a_{1:t}\sim \pi_{\theta},b \sim \mathcal{B}}}[-\log \pi_{\theta}\left(a_t|a_{1:t-1}\right)(r_{\bm{a}}-b)_{+}].
\end{equation}
Then, an empirical approximation of the Eq.~\ref{eq:sim} is:
\begin{equation}
    \label{eq:simappr}
    L_{\text{replay}} = \frac{1}{n}\sum_{k=1}^{n}\sum_{t=1}^{T}\nabla_{\theta}-\log \pi_{\theta}\left( a_t|a_{1:t-1}\right)(r_{\bm{a}}-b)_{+}, 
\end{equation}
where $n$ is the number of different child models that the controller samples in one batch and $T$ is the number of tokens.

Overall, the joint optimization process is specified in Algorithm~\ref{algo:optimization}, which consists of two phrases: the curiosity-guided search process and the self-imitation learning process. 
The optimal model with the best performance on the validation set is utilized for the outlier detection tasks.



\section{Experiments}
\label{sec:exp}
In this section, we conduct extensive experiments to answer the following four research questions.

\begin{itemize}[wide=0pt, leftmargin=\dimexpr\labelwidth + 2\labelsep\relax]

\item \textbf{Q1}: How effective is AutoOD compared with \textbf{state-of-the-art handcrafted algorithms}?

\item \textbf{Q2}: Whether or not the two key components of AutoOD, \textit{i.e.}, \textbf{curiosity-guided search and experience replay}, are effective in the search process?

\item \textbf{Q3}: Compared with \textbf{random search}, how effective is the proposed search strategy? 

\item \textbf{Q4}: Does AutoOD have the potential to be applied in \textbf{more complicated real-world applications}?

\end{itemize}

\subsection{Datasets and Tasks}
We evaluate \model on seven benchmark datasets for instance-level outlier sample detection and pixel-level defect region segmentation tasks. We also conduct a case study on the CAT~\cite{catset} dataset.

\begin{itemize}[wide=0pt, leftmargin=\dimexpr\labelwidth + 2\labelsep\relax]
    \item \textbf{MNIST~\cite{lecun1998gradient}}: An image dataset consists of handwritten digits. It has a training set of $60,000$ examples, and a test set of $10,000$ examples. 
    
    \item \textbf{Fashion-MNIST~\cite{xiao2017fashion}}: A MNIST-like dataset contains fashion product with  a training set of $60,000$ examples and a test set of $10,000$ examples. Each example is 
    a $28 \times 28$ grayscale image associated with a label from $10$ classes.
    
    \item \textbf{CIFAR-10~\cite{krizhevsky2009learning}}: A image dataset consists of $50,000$ training images and $10,000$ test images in $10$ different classes.   Each example is a $32 \times 32$ 3-channel image.
    
    
    \item \textbf{Tiny-ImageNet~\cite{deng2009imagenet}}: An image dataset consists of a subset of ImageNet images. It contains $10,000$ test images from $200$ different classes. We downsample each image to the size of $64 \times 64$.
    
    \item \textbf{MVTec-AD~\cite{bergmann2019mvtec}}: A benchmark dataset relates to industrial inspection in the application of anomaly detection. 
    It contains over $5000$ high-resolution images divided into fifteen categories in terms of different objects and textures. Each category comprises two parts: a training set of defect-free images, as well as a test set composed of defect-free images and the ones with various defects.
    We downsample each image to size $224 \times 224$.
    
    \item \textbf{CAT~\cite{catset}}: A cat dataset includes $10,000$ cat images. We downsample each image to size $224 \times 224$.
    
    \item \textbf{Gaussian Noise}:  A synthetic Gaussian noise dataset consists of $1,000$ random 2D images, where the value of each pixel is sampled from an i.i.d Gaussian distribution with mean $0.5$ and unit variance. We further clip each pixel into the range $[0,1]$.
    
    \item \textbf{Uniform Noise: } A synthetic uniform noise dataset consists of $1,000$ images, at which the value of each pixel is sampled from an i.i.d uniform distribution on $[0,1]$.
    
\end{itemize}

For the instance-level outlier sample detection task, we use four benchmark datasets (\textit{i.e.}, MNIST~\cite{lecun1998gradient}, Fashion-MNIST~\cite{xiao2017fashion}, CIFAR-10~\cite{krizhevsky2009learning},
and Tiny-ImageNet~\cite{deng2009imagenet}), and two synthetic noise datasets (\textit{i.e.}, Gaussian and Uniform). Synthetic noise datasets consist of $1,000$ random 2D images, where the value of each pixel is sampled from an i.i.d Gaussian distribution with mean $0.5$ and unit variance. We further clip each pixel into the range $[0,1]$, or an i.i.d uniform distribution on $[0,1]$. Different datasets contain different classes of images. We manually injected outlier samples (a.k.a. out-of-distribution samples), which consists of images randomly sampled from other datasets. For all six datasets, we train an outlier detection model on the training set, which only contains in-distribution samples, and use a validation set with out-of-distribution samples to guide the search, and another test set with out-of-distribution samples to evaluate the performance. The contamination ratio in the validation set and the test set are both $0.05$. The train/validation/test split ratio is $6:2:2$. Two state-of-the-art methods including MSP~\cite{hendrycks2016baseline} and ODIN~\cite{liang2017enhancing} are used as baselines.

For the pixel-level defect region segmentation task, we use a real-world dataset MVTec-AD~\cite{bergmann2019mvtec}. MVTec-AD contains high-resolution images with different objects and texture categories. Each category comprises a set of defect-free training images and a test set of images with various kinds of defects and images without defects. 
We train the model on the defect-free training set, and split the whole test set into two halves for validation and testing. Three state-of-the-art methods including AutoEncoder~\cite{bergmann2018improving}, AnoGAN~\cite{schlegl2017unsupervised}, and Feature Dictionary~\cite{napoletano2018anomaly} are used as baselines.


\subsection{Baselines}
We compare AutoOD with five state-of-the-art handcrafted algorithms and the random search strategy.

\begin{itemize}[wide=0pt, leftmargin=\dimexpr\labelwidth + 2\labelsep\relax]
    \item \textbf{MSP~\cite{hendrycks2016baseline}}: The softmax probability distribution is used to detect the anomalies in tasks of computer vision, natural language processing, and automatic speech. The outlier detection is performed based on the following assumption: the correctly classified examples have greater maximum softmax probabilities than those of erroneously classified and out-of-distribution examples.
    
    \item \textbf{ODIN~\cite{liang2017enhancing}}: The pre-trained neural networks is reused to detect the out-of-distribution images. ODIN separates the softmax probability distributions between in- and out-of-distribution instance, by using temperature scaling and adding small perturbations on the image data. 

    \item \textbf{AutoEncoder~\cite{bergmann2018improving}}: The structure of convolutional AutoEncoders is applied for unsupervised defect segmentation on image data. More specifically, it utilizes the loss function based on structural similarity, and successfully examines inter-dependencies between local image regions to reveal the defective regions.
    
     \item \textbf{AnoGAN~\cite{schlegl2017unsupervised}}: It is a deep convolutional generative adversarial network used to identify the anomalous image data. It learns a manifold of normal anatomical variability, and maps images to a latent space to estimate the anomaly scores. 
      
    \item \textbf{Feature Dictionary~\cite{napoletano2018anomaly}}: It applies the convolutional neural networks and self-similarity to detect and localize anomalies in image data. More specifically, the abnormality degree of each image region is obtained by estimating its similarity to a dictionary of anomaly-free subregions in a training set.
    
    \item \textbf{Random Search~\cite{bergstra2012random,li2019random}}: Instead of learning a policy to optimize the search progress, random search generates a neural architecture randomly at each step. It has been widely demonstrated that random search is a strong baseline hard to be surpassed in NAS. 
\end{itemize}


\begin{table*}[ht]
    \begin{subtable}[tc]{0.5\textwidth}
        \centering
        \caption{In-distribution dataset: MNIST}
            \begin{tabular}{@{\extracolsep{0.6pt}}lccc}
                \toprule 
                \begin{tabular}{@{}c@{}} OOD Dataset\end{tabular} &   \multicolumn{1}{c}{AUROC} & \multicolumn{1}{c}{AUPR In} &\multicolumn{1}{c}{AUPR Out} \\
                \hline 
 Fashion-MNIST  & \textbf{99.9}/97.9/97.9     & \textbf{99.9}/99.7/99.6 & \textbf{100}/90.5/91.0 \\
                notMNIST & \textbf{99.8}/97.2/98.2 & \textbf{99.8}/97.5/98.4  & \textbf{100}/97.4/98.0 \\
                CIFAR-10 & \textbf{99.9}/\textbf{99.9}/99.7 & 91.3/90.3/\textbf{99.9}  & 99.2/\textbf{99.9}/97.6 \\
                LSUN&\textbf{99.9}/\textbf{99.9}/99.8 &    99.9/96.8/\textbf{100}  & \textbf{99.9}/99.2/99.0 \\
                Tiny-ImageNet & \textbf{99.9}/99.4/99.6 & 99.8/99.6/\textbf{99.9} & \textbf{99.8}/96.8/97.5\\
                Gaussian & \textbf{99.9}/99.7/99.9 &\textbf{100}/99.8/\textbf{100} & \textbf{100}/99.7/\textbf{100}\\
                Uniform & \textbf{100}/99.9/\textbf{100}& \textbf{100}/99.9/\textbf{100}& \textbf{100}/99.9/\textbf{100}\\
                \hline \\
            \end{tabular}
    \end{subtable}%
            \begin{subtable}[tc]{.52\textwidth}
        \centering

        \caption{In-distribution dataset: Fashion-MNIST}
            \begin{tabular}{@{\extracolsep{0.6pt}}lccc}
                \toprule 
                \begin{tabular}{@{}c@{}}OOD Dataset\end{tabular} &  \multicolumn{1}{c}{AUROC} & \multicolumn{1}{c}{AUPR In} &\multicolumn{1}{c}{AUPR Out} \\
                \hline 
                MNIST & \textbf{99.9}/92.9/72.9     & \textbf{99.9}/82.8/91.6 & \textbf{99.9}/94.2/46.1 \\
                notMNIST  & \textbf{99.8}/96.9/80.2 & \textbf{99.1}/81.6/94.2 & \textbf{100}/99.4/57.7\\
                CIFAR-10   & \textbf{99.9}/88.2/96.6 & \textbf{99.5}/80.6/99.3 & \textbf{99.9}/97.2/80.4\\
                LSUN & \textbf{99.5}/89.7/96.0 &97.9/81.9/\textbf{99.2} & \textbf{99.9}/97.7/79.9\\
                Tiny-ImageNet  & \textbf{98.2}/87.7/95.5 & 90.7/80.4/\textbf{99.0} & \textbf{95.3}/97.1/82.5\\
                Gaussian  &\textbf{99.9}/97.2/89.6& \textbf{99.9}/82.24/98.0& \textbf{100}/99.5/48.2\\
                Uniform  & \textbf{99.9}/95.8/63.6 & \textbf{99.9}/82.9/91.4 & \textbf{99.9}/99.0/19.8\\
                \hline \\
            \end{tabular}
    \end{subtable}\\
    \begin{subtable}[tc]{0.5\textwidth}
        \centering

        \caption{In-distribution dataset: CIFAR-10}
            \begin{tabular}{@{\extracolsep{0.6pt}}lccc}
                \toprule 
                 \begin{tabular}{@{}c@{}}OOD Dataset\end{tabular} &  \multicolumn{1}{c}{AUROC} & \multicolumn{1}{c}{AUPR In} &\multicolumn{1}{c}{AUPR Out} \\
                \hline 
                MNIST & \textbf{100}/98.4/99.9     & \textbf{100}/99.4/\textbf{100}& \textbf{100}/89.4/99.4 \\
                Fashion-MNIST & \textbf{99.6}/98.2/99.4 & 98.1/96.1/\textbf{99.9} & \textbf{99.9}/98.8/97.3\\
                notMNIST & \textbf{99.9}/96.8/98.1 & \textbf{99.4}/99.0/99.2 & \textbf{100}/89.4/90.2\\
                LSUN&  80.0/75.2/\textbf{85.6} & 81.2/73.1/\textbf{83.5} & \textbf{85.8}/73.3/85.1\\
                Tiny-ImageNet  & \textbf{84.0}/72.6/81.6 & \textbf{87.5}/73.5/76.9 & \textbf{87.8}/80.6/84.8\\
                Gaussian &\textbf{99.9}/86.3/98.8& \textbf{99.9}/90.5/99.1& \textbf{99.3}/77.0/97.9\\
                Uniform  & \textbf{99.9}/86.4/99.0 & \textbf{99.9}/90.2/99.2 & \textbf{99.9}/78.6/98.6\\
                \hline \\
            \end{tabular}
    \end{subtable}%
    \begin{subtable}[tc]{.5\textwidth}
        \centering

        \caption{In-distribution dataset: Tiny-ImageNet}
            \begin{tabular}{@{\extracolsep{0.6pt}}lccc}
                \toprule 
                \begin{tabular}{@{}c@{}}OOD Dataset\end{tabular}  & \multicolumn{1}{c}{AUROC} & \multicolumn{1}{c}{AUPR In} &\multicolumn{1}{c}{AUPR Out} \\
                \hline 
                MNIST & \textbf{100}/99.8/94.8     & \textbf{100}/98.2/98.9 & \textbf{100}/98.2/79.9 \\
                Fashion-MNIST & \textbf{99.7}/70.4/73.8& \textbf{98.6}/88.4/92.6 & \textbf{100}/85.5/39.5\\
                notMNIST  & \textbf{99.9}/80.6/82.7 & \textbf{99.7}/90.4/95.2
                & \textbf{100}/85.8/56.0\\
                CIFAR-10  & \textbf{86.7}/82.9/58.0 & \textbf{95.8}/75.3/85.7 & \textbf{89.1}/75.3/28.2\\
                LSUN & \textbf{88.6}/74.2/55.6 & 92.7/\textbf{99.5}/86.7& 93.9/\textbf{99.5}/18.7\\
                Gaussian & \textbf{99.9}/97.0/95.4 &\textbf{100}/98.0/99.0 & \textbf{99.9}/94.8/80.5\\
                Uniform & \textbf{100}/96.0/87.5 & \textbf{100}/97.4/96.8 & \textbf{100}/99.3/62.8 \\

                \hline \\
            \end{tabular}
    \end{subtable}%
    \caption{Performance comparison on instance-level outlier sample detection. The results from AutoOD and the two baselines MSP~\cite{hendrycks2016baseline} and ODIN~\cite{liang2017enhancing} are listed as AutoOD/MSP/ODIN. OOD: Out-of-distribution.}
    \label{tab:instancedetection}
\end{table*}

\begin{table}[t]
\centering
\begin{tabular}{@{}cccccc@{}}
\toprule
\multicolumn{2}{c}{Category} & \multicolumn{1}{c}{Our Model}                        & \multicolumn{1}{c}{AutoEncoder}                      & \multicolumn{1}{c}{AnoGAN}                           & \multicolumn{1}{c}{\begin{tabular}[c]{@{}c@{}}Feature\\ Dictionary\end{tabular}} \\
\midrule
\multirow{5}{*}{\rotatebox{90}{Textures}}   & Carpet        &  \textbf{0.69}  /    \textbf{0.92}                      & 0.38                     / 0.59                       & 0.34                     / 0.54                       & 0.20                                  / 0.72                                                \\
                              & Grid          &      \textbf{0.89}                  /      \textbf{0.94}                      & 0.83                     / 0.90                       & 0.04                    / 0.58                       & 0.02                                  / 0.59                                                                       \\
                              & Leather       &  \textbf{0.81}   /          \textbf{0.92}                  & 0.67                     / 0.75                       & 0.34                    / 0.64                       & 0.74                                  / 0.87                                          \\
                              & Tile          &   \textbf{0.26}                       /       \textbf{0.94}                & 0.23                     / 0.51                       & 0.08                     / 0.50                       & 0.14                                   / 0.73                                                                          \\
                              & Wood          &       \textbf{0.47}                  /  \textbf{0.97}                         & 0.29                     / 0.73                       & 0.14                     / 0.62                       & \textbf{0.47}                                  / 0.91                                                                            \\ \midrule
\multirow{10}{*}{\rotatebox{90}{Objects}}     & Bottle        &      \textbf{0.33}                  /          \textbf{0.93}                  & 0.22                    / 0.86                       & 0.05                     / 0.86                       & 0.07                                   / 0.78                                                               \\
                              & Cable         &             \textbf{0.17}             /      \textbf{0.87}                    & 0.05                     / 0.86                       & 0.01                    / 0.78                       & 0.13                                   / 0.79  \\                                   & Capsule       &             \textbf{0.16}            /   \textbf{0.95}                        & 0.11                     / 0.88                       & 0.04                    / 0.84                       & 0.00                                   / 0.84                                                                  \\
                              & Hazelnut      &          \textbf{0.46}               /     \textbf{0.97}                     & 0.41                     / 0.95                       & 0.02                     / 0.87                       & 0.00                                   / 0.72                                                                      \\
                              & Metal Nut     &               \textbf{0.30}           /     \textbf{0.88}                       & 0.26                   / 0.86                       & 0.00                     / 0.76                       & 0.13                                  / 0.82                                                                  \\
                              & Pill          &            \textbf{0.30}              / \textbf{0.92}                          & 0.25                     / 0.85                       & 0.17                     / 0.87                       & 0.00                                 / 0.68                                                                     \\
                              & Screw         &              \textbf{0.34}            / \textbf{0.96}                           & \textbf{0.34}                     /  \textbf{0.96}                       & 0.01                     / 0.80                       & 0.00                                   / 0.87                                                                     \\
                              & Toothbrush    &              \textbf{0.60}            /    \textbf{0.90}                       & 0.51                    / 0.83                       & 0.07                     / \textbf{0.90}                       & 0.00                                   / 0.77                                                                      \\
                              & Transistor    &               \textbf{0.23}           /           \textbf{0.96}                 & 0.22                   / 0.86                       & 0.08                     / 0.80                       & 0.03                                / 0.66                                                                       \\
                              & Zipper        & \textbf{0.20}  /    \textbf{0.88}                        & 0.13                    / 0.77                       & 0.01                     / 0.78                       & 0.00                                  / 0.76                                                                     \\  \hline
\end{tabular}
    \caption{Performance comparison on pixel-level defect region segmentation. The results of RPRO and AUROC are listed as RPRO/AUROC. Baseline results are directly collected from the original paper~\cite{bergmann2019mvtec}.} 
    \label{tab:pixelleveldetection}. 
\vspace{-5pt}
\end{table}

\subsection{Experiment Setup}
We train the child models on the training set under the outlier-free settings, and update the controller on the validation set via the reward signal. The controller RNN is a two-layer LSTM with $50$ hidden units on each layer. It is trained with the ADAM optimizer with a learning rate of $3.5e-4$. Weights are initialized uniformly in $[-0.1,0.1]$. The search process is conducted for a total of $500$ epochs. The size of the self-imitation buffer is $10$. We use a Tanh constant of $2.5$ and a sample temperature of $5$ to the hidden output of the RNN controller. We train the child models by utilizing a batch size of $64$ and a momentum of $0.9$ with the ADAM optimizer. The learning rate starts at $0.1$, and is dropped by a factor of $10$ at $50\%$ and $75\%$ of the training progress, respectively.

\subsection{Evaluation Metrics}
We adopt the following metrics to measure the effectiveness: 
\begin{itemize}[wide=0pt, leftmargin=\dimexpr\labelwidth + 2\labelsep\relax]
    
 \item \textbf{AUROC}~\cite{davis2006relationship} is the Area Under the Receiver Operating Characteristic curve, which is a threshold-independent metric ~\cite{davis2006relationship}. The ROC curve depicts the relationship between TPR and FPR. The AUROC can be interpreted as the probability that a positive example is assigned a higher detection score than a negative example~\cite{fawcett2006introduction}.

 \item \textbf{AUPR}~\cite{manning1999foundations} is the Area under the Precision-Recall curve, which is another threshold-independent metric~\cite{manning1999foundations,saito2015precision}. The PR curve is a graph showing the precision=TP/(TP+FP) and recall=TP/(TP+FN) against each other. The metrics \textbf{AUPR-In} and \textbf{AUPR-Out} denote the area under the precision-recall curve, where positive samples and negative samples are specified as positives, respectively.

 \item \textbf{RPRO}~\cite{bergmann2019mvtec} stands for the relative per-region overlap. It denotes the pixel-wise overlap rate of the segmentations with the ground truth.
\end{itemize}

\subsection{Results}
\subsubsection{Performance on Out-of-distribution Sample Detection} 
To answer the research question \textbf{Q1}, we compare AutoOD with the state-of-the-art handcrafted algorithms for the instance-level outlier sample detection task using metrics AUROC, AUPR-In and AUPR-Out. Considering the automated search framework of AutoOD, we represent its performance by the best model found during the search process. In these experiments, we follow the setting in~\cite{hendrycks2016baseline,devries2018learning}: Each model is trained on individual dataset $\mathcal{D}_{in}$, which is taken from MNIST, Fashion-MNIST, CIFAR-10, and Tiny-ImageNet, respectively. At test time, the test images from $\mathcal{D}_{in}$ dataset can be viewed as the in-distribution (positive) samples. We sample out-of-distribution (negative) images from another real-world or synthetic noise dataset, after down-sampling/up-sampling and reshaping their sizes as the same as $\mathcal{D}_{in}$. 


As can be seen from Table~\ref{tab:instancedetection}, in most of the test cases, the models discovered by AutoOD consistently outperform the handcrafted out-of-distribution detection methods with pre-trained models (ODIN~\cite{liang2017enhancing}) and without pre-trained models (MSP~\cite{hendrycks2016baseline}).
It indicates that AutoOD could achieve higher performance in accuracy, precision, and recall simultaneously, with a more precise detection rate and fewer nuisance alarms. 

\subsubsection{Performance on Defect Sample Detection}

To further answer question \textbf{Q1}, we test AutoOD on the pixel-level defect region segmentation task. The results from Table \ref{tab:pixelleveldetection} show that AutoOD consistently outperforms the baseline methods by a large margin in terms of AUROC and RPRO. The higher AUROC demonstrates that the model found by AutoOD precisely detects images with defect sections out of the positive samples. The results also show that AutoOD has a better performance in RPRO. This indicates the search process of AutoOD helps the model to locate and represent the outlier regions in negative images.

\begin{table*}[ht]
\centering
\begin{tabular}{@{}lccccc@{}}
\toprule
Actions               & MNIST & Fashion-MNIST & CIFAR-10 & Tiny-ImageNet & MVTec-AD \\ \midrule
Definition hypothesis &   reconstruction    &                   cluster     &         centroid      &     reconstruction     &     reconstruction         \\
Distance measurement  &   $l_{1}$    &                  $l_{2,1}$        &       $l_{2,1}$       &        $l_{2,1}$     &       $l_{2,1}+$ SSIM           \\
\textit{Layer-1}               &       &               &          &               &          \\
\quad Output-channel        &    32   &                  16     &           16    &     16        &            32  \\
\quad Convolution kernel    &     $5 \times 5 $   &                  $1 \times 1 $     &      $1 \times 1 $           &    $5 \times 5 $            &     $1 \times 1 $           \\
\quad Pooling type          &  mean     &                    average    &       average        &     average      &          average   \\
\quad Pooling kernel        &    $1 \times 1 $      &                    $1 \times 1 $   &        $1 \times 1 $        &     $1 \times 1 $           &       $3 \times 3 $          \\
\quad Normalization type    &  no     &                     no   &         no        &      no     &           no  \\
\quad Activation function   &  ReLU     &                   Sigmoid     &       Sigmoid        &    Linear      &      LeakyReLU        \\
\textit{Layer-2}                &       &               &          &               &          \\
\quad Output-channel        &   8    &                32        &          32     &         32      &           32  \\
\quad Convolution kernel    &    $3 \times 3 $    &                    $5 \times 5 $       &       $5 \times 5 $           &      $3 \times 3 $       &         $5 \times 5 $      \\
\quad Pooling type          &  average     &                  mean      &     mean          &       mean    &   mean          \\
\quad Pooling kernel        &    $1 \times 1 $      &                $7 \times 7$       &           $7 \times 7$         &     $7 \times 7$        &    $5 \times 5 $             \\
\quad Normalization type    &   no    &                    no    &          no       &     no    &     batch          \\
\quad Activation function   &  ELU     &                 ReLU6     &        Sigmoid       &      Softplus    &         Tanh     \\
\textit{Layer-3}                &       &               &          &               &          \\
\quad Output-channel        &   8    &                   16     &        16       &     16       &           16  \\
\quad Convolution kernel    &   $7 \times 7$      &                  $1 \times 1 $        &       $1 \times 1 $         &   $7\times 7 $        &    $1 \times 1 $             \\
\quad Pooling type          &   average    &                  average     &        average       &    average     &     mean        \\
\quad Pooling kernel        &     $5 \times 5 $     &                  $1 \times 1 $        &          $1 \times 1 $             &    $1 \times 1 $        &     $5 \times 5 $            \\
\quad Normalization type    &  no     &                    instance    &      instance         &     instance    &        instance       \\
\quad Activation function   & ReLU6      &                 LeakyReLU     &      Sigmoid         &      LeakyReLU   &         Tanh      \\
 \bottomrule
\end{tabular}
\caption{The architectures discovered by AutoOD for MNIST, Fashion-MNIST, CIFAR-10, Tiny-ImageNet, and MVTec-AD.}
\label{tab:bestarchitecture}
\end{table*}

Table~\ref{tab:bestarchitecture} illustrates the best architectures discovered by AutoOD on both instance-level outlier sample detection and pixel-level defect region segmentation tasks.

\subsubsection{Effectiveness of Curiosity-guided Search}
To qualitatively evaluate the effectiveness of the curiosity-guided search for research question \textbf{Q2}, we perform ablation and hyperparameter analysis on Fashion-MNIST dataset with samples from CIFAR-10 as outliers. Specifically, we control the hyperparameter $\eta$ in Eq.~\ref{eq:reward} for illustration. Note that mathematical expression of $\eta=0$ represents that there is no exploration. 
From Table~\ref{tab:ablation} (a) we can observe that: (1) The absence of exploration would negatively impact the final performance. The AUROC after 200 epochs could drop $1.9\%$. (2) The curiosity guided explorations help the controller find the optimal model faster. The better performance could be achieved in the $20$-th, $100$-th epochs when the controller has a larger weight $\eta$ on explorations. This indicates that curiosity-guided search is a promising way for exploring more unseen spaces. 
(3) Comparing AUROC between $\eta=0.01$ and $\eta=0.1$, we observe that there is no significant increase in the final performance after 200 epochs. This indicates that a higher rate of explorations can not always guarantee a higher performance. (4) If we treat the performance of the searched result as Gaussian distributions, the standard deviations of the AutoOD's performance keep increasing when $\eta$ increases. This validates that the curiosity-guided search strategy increases the opportunity for the controller to generate child models in a more diverse way. 

\subsubsection{Effectiveness of Experience Replay}
To further answer the question \textbf{Q2}, we evaluate the effectiveness of the experience replay buffers, by altering the size of the replay buffers $\mathcal{B}$ in Eq.~\ref{eq:simappr}. Corresponding results are reported in Table \ref{tab:ablation} (b). The results indicate that the increase of the buffer size could enhance model performance after 200 epochs. We also observe that the size of the buffer is sensitive to the final performance, as better performance would be achieved in the 20th, 100th epoch with a larger buffer size. This indicates that self-imitation learning based experience replay is useful in the search process. Larger buffer size brings benefits to exploit past good experiences.

\begin{table}[t]
    \centering
   \begin{subtable}[t]{.5\textwidth}
       \centering

        \caption{Curiosity-guided Search}
\begin{tabular}{@{}cccccc@{}}
\toprule
          & $\textrm{AUROC}_{20}$& $\textrm{AUROC}_{100}$  & $\textrm{AUROC}_{200}$ & $\textrm{mean}_{200}$ & $\textrm{std}_{200}$ \\ \midrule

$\eta$=$0$ &85.27&     96.23 & 96.51 &   90.01  &  0.093       \\
$\eta$=$0.01$ &85.46&      96.54   & 98.50 &  91.20     &    0.097       \\
$\eta$=$0.1$ & 85.46 &    96.93& 98.41    & 91.84      &0.131         \\ \hline
\end{tabular}
    \end{subtable}\\
    \vspace{8pt}
    \begin{subtable}[t]{.5\textwidth}
        \centering
        \caption{Experience Replay Buffer}
\begin{tabular}{@{}cccc@{}}

\toprule
               &$\textrm{AUROC}_{20}$ & $\textrm{AUROC}_{100}$ & $\textrm{AUROC}_{200}$  \\ \midrule
   
no buffer      &    85.43 &     96.57  &  98.00    \\
buffer size=$5$  & 88.05     &  97.12     &    98.04   \\
buffer size=$10$ &  87.44    & 97.70    &98.12      \\  \hline
\end{tabular}
    \end{subtable}\\
    \caption{Ablations and parameter analysis on Fashion-MNIST (In-distribution: normal data) and CIFAR-10 (Out-of-distribution: outliers). We report the performance of AutoOD under different search strategies and hyper-parameter settings, with $20_{th}$, $100_{th}$, and $200_{th}$ iterations, respectively.}
\label{tab:ablation}
\end{table}

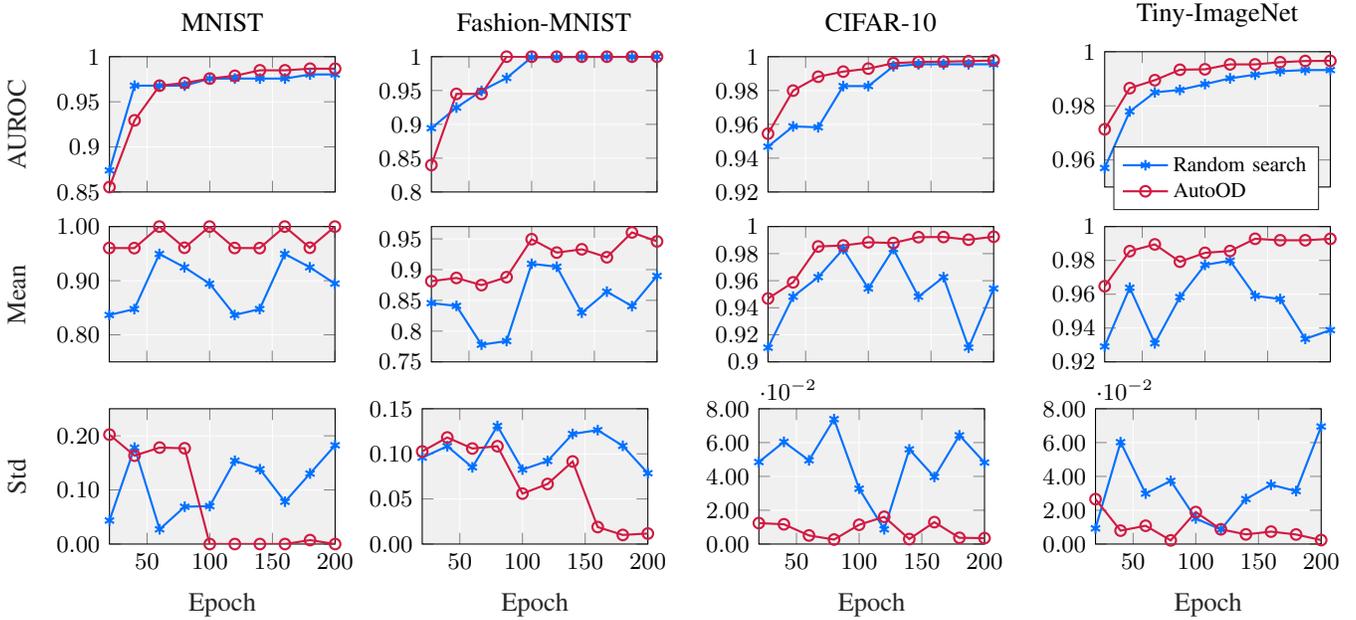
\begin{figure*}[t]
\captionsetup{font=small}
\captionsetup[subfigure]{justification=centering}
\begin{subfigure}[b]{0.24\textwidth}
 \centering 
%
%
\definecolor{mycolor1}{rgb}{0.0, 0.44, 1.0}%
\definecolor{mycolor2}{rgb}{0.65000,0.22500,0.09800}%
\definecolor{mycolor3}{rgb}{0.82, 0.1, 0.26}%

\begin{tikzpicture}

\begin{axis}[%
xticklabel=\empty,
label style={font=\small},
tick label style={font=\small},
width=\figurewidth,
height=\figureheight,
at={(0\figurewidth,0\figureheight)},
scale only axis,
xmin=20,
xmax=200,
xlabel style={font=\color{white!15!black}},
ymin=0.85,
ymax=1,
grid, 
grid style={line width=.15pt, draw=white!0}, 
ylabel style={font=\color{white!15!black}},
ylabel={AUROC},
title={MNIST},
axis background/.style={fill=white},
legend style={legend cell align=left, align=left, draw=white!15!black, nodes={scale=0.8}, at={(0.95, 0.3)}},
axis background/.style={fill=gray!12} 
]
\addplot [color=mycolor1, mark=asterisk, mark options={solid, mycolor1}, thick]
  table[row sep=crcr]{%
20  0.8739594587\\
40  0.9678856229\\
60  0.9678856229\\
80  0.9678856229\\
100 0.9756804308\\
120 0.9757462204\\
140 0.9757462204\\
160 0.9757462204\\
180 0.9802941636\\
200 0.9802941636\\
};

\addplot [color=mycolor3, mark=o, mark options={solid, mycolor3}, thick]
  table[row sep=crcr]{%
20  0.8554338822\\
40  0.92940006\\
60  0.9678879477\\
80  0.9708924934\\
100 0.97576354\\
120 0.9787757616\\
140 0.9848797919\\
160 0.9848797919\\
180 0.9866794263\\
200 0.9866794263\\
};
\end{axis}
\end{tikzpicture}%
\end{subfigure}%
\hspace{7pt}
\begin{subfigure}[b]{0.24\textwidth}
 \centering 
%
%
\definecolor{mycolor1}{rgb}{0.0, 0.44, 1.0}%
\definecolor{mycolor2}{rgb}{0.65000,0.22500,0.09800}%
\definecolor{mycolor3}{rgb}{0.82, 0.1, 0.26}%

\begin{tikzpicture}

\begin{axis}[%
xticklabel=\empty,
label style={font=\small},
tick label style={font=\small},
width=\figurewidth,
height=\figureheight,
at={(0\figurewidth,0\figureheight)},
scale only axis,
xmin=20,
xmax=200,
xlabel style={font=\color{white!15!black}},
ymin=0.8,
ymax=1,
grid, 
grid style={line width=.15pt, draw=white!0}, 
ylabel style={font=\color{white!15!black}},
title={Fashion-MNIST},
axis background/.style={fill=white},
legend style={legend cell align=left, align=left, draw=white!15!black, nodes={scale=0.8}, at={(0.95, 0.3)}},
axis background/.style={fill=gray!12} 
]
\addplot [color=mycolor1, mark=asterisk, mark options={solid, mycolor1}, thick]
  table[row sep=crcr]{%
20  0.8942665628\\
40  0.92475898\\
60  0.94944621\\
80  0.9685442235\\
100 0.9991186932\\
120 0.9991186932\\
140 0.9996671445\\
160 0.9996671445\\
180 0.9996671445\\
200 0.9999170507\\
};

\addplot [color=mycolor3, mark=o, mark options={solid, mycolor3}, thick]
  table[row sep=crcr]{%
20  0.8398022231\\
40  0.9449847866\\
60  0.9449847866\\
80  1\\
100 1\\
120 1\\
140 1\\
160 1\\
180 1\\
200 1\\
};
\end{axis}
\end{tikzpicture}%
\end{subfigure}
\begin{subfigure}[b]{0.24\textwidth}
 \centering 
%
%
\definecolor{mycolor1}{rgb}{0.0, 0.44, 1.0}%
\definecolor{mycolor2}{rgb}{0.65000,0.22500,0.09800}%
\definecolor{mycolor3}{rgb}{0.82, 0.1, 0.26}%

\begin{tikzpicture}

\begin{axis}[%
xticklabel=\empty,
label style={font=\small},
tick label style={font=\small},
width=\figurewidth,
height=\figureheight,
at={(0\figurewidth,0\figureheight)},
scale only axis,
xmin=20,
xmax=200,
xlabel style={font=\color{white!15!black}},
ymin=0.92,
ymax=1,
grid, 
grid style={line width=.15pt, draw=white!0}, 
ylabel style={font=\color{white!15!black}},
title={CIFAR-10},
axis background/.style={fill=white},
legend style={legend cell align=left, align=left, draw=white!15!black, nodes={scale=0.8}, at={(0.95, 0.3)}},
axis background/.style={fill=gray!12} 
]
\addplot [color=mycolor1, mark=asterisk, mark options={solid, mycolor1}, thick]
  table[row sep=crcr]{%
20  0.9468334781\\
40  0.9587778466\\
60  0.9582414174\\
80  0.9826204324\\
100 0.9826204324\\
120 0.9943258172\\
140 0.9955033094\\
160 0.9955033094\\
180 0.9955033094\\
200 0.9955033094\\
};

\addplot [color=mycolor3, mark=o, mark options={solid, mycolor3}, thick]
  table[row sep=crcr]{%
20  0.9544958741\\
40  0.9799304105\\
60  0.9881535223\\
80  0.9911777695\\
100 0.9929317132\\
120 0.9961757264\\
140 0.9968734851\\
160 0.9970704907\\
180 0.9973556389\\
200 0.9977609813\\
};
\end{axis}
\end{tikzpicture}%
\end{subfigure}
\begin{subfigure}[b]{0.24\textwidth}
 \centering 
%
%
\definecolor{mycolor1}{rgb}{0.0, 0.44, 1.0}%
\definecolor{mycolor2}{rgb}{0.65000,0.22500,0.09800}%
\definecolor{mycolor3}{rgb}{0.82, 0.1, 0.26}%

\begin{tikzpicture}

\begin{axis}[%
xticklabel=\empty,
label style={font=\small},
tick label style={font=\small},
width=\figurewidth,
height=\figureheight,
at={(0\figurewidth,0\figureheight)},
scale only axis,
xmin=20,
xmax=200,
xlabel style={font=\color{white!15!black}},
ymin=0.95,
ymax=1,
grid, 
grid style={line width=.15pt, draw=white!0}, 
ylabel style={font=\color{white!15!black}},
title={Tiny-ImageNet},
axis background/.style={fill=white},
legend style={legend cell align=left, align=left, draw=white!15!black, nodes={scale=0.8}, at={(0.95, 0.3)}},
axis background/.style={fill=gray!12} 
]
\addplot [color=mycolor1, mark=asterisk, mark options={solid, mycolor1}, thick]
  table[row sep=crcr]{%
20  0.9570163263\\
40  0.9779428189\\
60  0.9849807826\\
80  0.9858877532\\
100 0.9880062293\\
120 0.9901605239\\
140 0.9915179843\\
160 0.9928021408\\
180 0.9932202381\\
200 0.9932202381\\
};\addlegendentry{Random search}

\addplot [color=mycolor3, mark=o, mark options={solid, mycolor3}, thick]
  table[row sep=crcr]{%
20  0.9713346493\\
40  0.9865375399\\
60  0.9894795977\\
80  0.9933356661\\
100 0.9934979242\\
120 0.9953149181\\
140 0.9953149181\\
160 0.9961132611\\
180 0.9965682489\\
200 0.9966285957\\
};\addlegendentry{AutoOD}
\end{axis}
\end{tikzpicture}%
\end{subfigure}


\begin{subfigure}[b]{0.24\textwidth}
 \centering 
%
%
\definecolor{mycolor1}{rgb}{0.0, 0.44, 1.0}%
\definecolor{mycolor2}{rgb}{0.65000,0.22500,0.09800}%
\definecolor{mycolor3}{rgb}{0.82, 0.1, 0.26}%

\begin{tikzpicture}

\begin{axis}[%
xticklabel=\empty,
label style={font=\small},
tick label style={font=\small},
y tick label style={/pgf/number format/.cd, fixed,fixed zerofill,precision=2,/tikz/.cd},
width=\figurewidth,
height=\figureheight,
at={(0\figurewidth,0\figureheight)},
scale only axis,
xmin=20,
xmax=200,
xlabel style={font=\color{white!15!black}},
ymin=0.75,
ymax=1,
grid, 
grid style={line width=.15pt, draw=white!0}, 
ylabel style={font=\color{white!15!black}},
ylabel={Mean},
axis background/.style={fill=white},
legend style={legend cell align=left, align=left, draw=white!15!black, nodes={scale=0.8}, at={(0.95, 0.3)}},
axis background/.style={fill=gray!12} 
]
\addplot [color=mycolor1, mark=asterisk, mark options={solid, mycolor1}, thick]
  table[row sep=crcr]{%
20  0.83660794\\
40  0.84755774\\
60  0.94944621\\
80  0.92475898\\
100 0.89462523\\
120 0.83660794\\
140 0.84755774\\
160 0.94944621\\
180 0.92475898\\
200 0.89462523\\
};

\addplot [color=mycolor3, mark=o, mark options={solid, mycolor3}, thick]
  table[row sep=crcr]{%
20  0.9604047\\
40  0.96018873\\
60  1\\
80  0.96083973\\
100 0.99999706\\
120 0.9604047\\
140 0.96018873\\
160 1\\
180 0.96083973\\
200 0.99999706\\
};
\end{axis}
\end{tikzpicture}%
\end{subfigure}%
\hspace{7pt}
\begin{subfigure}[b]{0.24\textwidth}
 \centering 
%
%
\definecolor{mycolor1}{rgb}{0.0, 0.44, 1.0}%
\definecolor{mycolor2}{rgb}{0.65000,0.22500,0.09800}%
\definecolor{mycolor3}{rgb}{0.82, 0.1, 0.26}%

\begin{tikzpicture}

\begin{axis}[%
xticklabel=\empty,
label style={font=\small},
tick label style={font=\small},
width=\figurewidth,
height=\figureheight,
at={(0\figurewidth,0\figureheight)},
scale only axis,
xmin=20,
xmax=200,
xlabel style={font=\color{white!15!black}},
ymin=0.75,
ymax=0.97,
grid, 
grid style={line width=.15pt, draw=white!0}, 
ylabel style={font=\color{white!15!black}},
axis background/.style={fill=white},
legend style={legend cell align=left, align=left, draw=white!15!black, nodes={scale=0.8}, at={(0.95, 0.3)}},
axis background/.style={fill=gray!12} 
]
\addplot [color=mycolor1, mark=asterisk, mark options={solid, mycolor1}, thick]
  table[row sep=crcr]{%
20  0.84546583\\
40  0.84108642\\
60  0.77793686\\
80  0.78381318\\
100 0.90950664\\
120 0.90472342\\
140 0.82999834\\
160 0.86419313\\
180 0.84070725\\
200 0.88975895\\
};

\addplot [color=mycolor3, mark=o, mark options={solid, mycolor3}, thick]
  table[row sep=crcr]{%
20  0.88119181\\
40  0.88646448\\
60  0.8747805\\
80  0.88769532\\
100 0.949423\\
120 0.92767524\\
140 0.93304307\\
160 0.92007248\\
180 0.96049848\\
200 0.94591242\\
};
\end{axis}
\end{tikzpicture}%
\end{subfigure}
\begin{subfigure}[b]{0.24\textwidth}
 \centering 
%
%
\definecolor{mycolor1}{rgb}{0.0, 0.44, 1.0}%
\definecolor{mycolor2}{rgb}{0.65000,0.22500,0.09800}%
\definecolor{mycolor3}{rgb}{0.82, 0.1, 0.26}%

\begin{tikzpicture}

\begin{axis}[%
xticklabel=\empty,
label style={font=\small},
tick label style={font=\small},
width=\figurewidth,
height=\figureheight,
at={(0\figurewidth,0\figureheight)},
scale only axis,
xmin=20,
xmax=200,
xlabel style={font=\color{white!15!black}},
ymin=0.9,
ymax=1,
grid, 
grid style={line width=.15pt, draw=white!0}, 
ylabel style={font=\color{white!15!black}},
axis background/.style={fill=white},
legend style={legend cell align=left, align=left, draw=white!15!black, nodes={scale=0.8}, at={(0.95, 0.3)}},
axis background/.style={fill=gray!12} 
]
\addplot [color=mycolor1, mark=asterisk, mark options={solid, mycolor1}, thick]
  table[row sep=crcr]{%
20  0.91049202\\
40  0.94820253\\
60  0.962608\\
80  0.98340898\\
100 0.95422241\\
120 0.98340898\\
140 0.94820253\\
160 0.962608\\
180 0.91049202\\
200 0.95422241\\
};

\addplot [color=mycolor3, mark=o, mark options={solid, mycolor3}, thick]
  table[row sep=crcr]{%
20  0.9468334781\\
40  0.9587778466\\
60  0.98530583\\
80  0.98600526\\
100 0.9883579\\
120 0.98785999\\
140 0.99216558\\
160 0.99228648\\
180 0.99028579\\
200 0.99249996\\
};
\end{axis}
\end{tikzpicture}%
\end{subfigure}
\begin{subfigure}[b]{0.24\textwidth}
 \centering 
%
%
\definecolor{mycolor1}{rgb}{0.0, 0.44, 1.0}%
\definecolor{mycolor2}{rgb}{0.65000,0.22500,0.09800}%
\definecolor{mycolor3}{rgb}{0.82, 0.1, 0.26}%

\begin{tikzpicture}

\begin{axis}[%
xticklabel=\empty,
label style={font=\small},
tick label style={font=\small},
width=\figurewidth,
height=\figureheight,
at={(0\figurewidth,0\figureheight)},
scale only axis,
xmin=20,
xmax=200,
xlabel style={font=\color{white!15!black}},
ymin=0.92,
ymax=1,
grid, 
grid style={line width=.15pt, draw=white!0}, 
ylabel style={font=\color{white!15!black}},
axis background/.style={fill=white},
legend style={legend cell align=left, align=left, draw=white!15!black, nodes={scale=0.8}, at={(0.95, 0.3)}},
axis background/.style={fill=gray!12} 
]
\addplot [color=mycolor1, mark=asterisk, mark options={solid, mycolor1}, thick]
  table[row sep=crcr]{%
20  0.92912691\\
40  0.96386117\\
60  0.93096688\\
80  0.95830473\\
100 0.97734994\\
120 0.97986732\\
140 0.959053\\
160 0.95718693\\
180 0.93359093\\
200 0.93878112\\
};

\addplot [color=mycolor3, mark=o, mark options={solid, mycolor3}, thick]
  table[row sep=crcr]{%
20  0.96474227\\
40  0.98547422\\
60  0.98946631\\
80  0.97922754\\
100 0.98447155\\
120 0.98555442\\
140 0.9927839\\
160 0.99191281\\
180 0.99191281\\
200 0.99275707\\
};
\end{axis}
\end{tikzpicture}%
\end{subfigure}

\begin{subfigure}[b]{0.24\textwidth}
 \centering 
%
%
\definecolor{mycolor1}{rgb}{0.0, 0.44, 1.0}%
\definecolor{mycolor2}{rgb}{0.65000,0.22500,0.09800}%
\definecolor{mycolor3}{rgb}{0.82, 0.1, 0.26}%

\begin{tikzpicture}

\begin{axis}[%
label style={font=\small},
tick label style={font=\small},
y tick label style={/pgf/number format/.cd, fixed,fixed zerofill,precision=2,/tikz/.cd},
width=\figurewidth,
height=\figureheight,
at={(0\figurewidth,0\figureheight)},
scale only axis,
xmin=20,
xmax=200,
xlabel style={font=\color{white!15!black}},
xlabel={Epoch},
ymin=0,
ymax=0.25,
grid, 
grid style={line width=.15pt, draw=white!0}, 
ylabel style={font=\color{white!15!black}},
ylabel={Std},
axis background/.style={fill=white},
legend style={legend cell align=left, align=left, draw=white!15!black, nodes={scale=0.8}, at={(0.95, 0.3)}},
axis background/.style={fill=gray!12} 
]
\addplot [color=mycolor1, mark=asterisk, mark options={solid, mycolor1}, thick]
  table[row sep=crcr]{%
20  0.04352645\\
40  0.178209188\\
60  0.02726623\\
80  0.06897495\\
100 0.07007477\\
120 0.15399112\\
140 0.13832036\\
160 0.07817056\\
180 0.13032127\\
200 0.18251445\\
};

\addplot [color=mycolor3, mark=o, mark options={solid, mycolor3}, thick]
  table[row sep=crcr]{%
20  0.202197917\\
40  0.163371526\\
60  0.178209188\\
80  0.176956593\\
100 0\\
120 0\\
140 0\\
160 0\\
180 0.00732358025\\
200 0.0000123365442\\
};
\end{axis}
\end{tikzpicture}%
\end{subfigure}%
\hspace{7pt}
\begin{subfigure}[b]{0.24\textwidth}
 \centering 
%
%
\definecolor{mycolor1}{rgb}{0.0, 0.44, 1.0}%
\definecolor{mycolor2}{rgb}{0.65000,0.22500,0.09800}%
\definecolor{mycolor3}{rgb}{0.82, 0.1, 0.26}%

\begin{tikzpicture}

\begin{axis}[%
label style={font=\small},
tick label style={font=\small},
y tick label style={/pgf/number format/.cd, fixed,fixed zerofill,precision=2},
width=\figurewidth,
height=\figureheight,
at={(0\figurewidth,0\figureheight)},
scale only axis,
xmin=20,
xmax=200,
xlabel style={font=\color{white!15!black}},
xlabel={Epoch},
ymin=0.,
ymax=0.15,
grid, 
grid style={line width=.15pt, draw=white!0}, 
ylabel style={font=\color{white!15!black}},
axis background/.style={fill=white},
legend style={legend cell align=left, align=left, draw=white!15!black, nodes={scale=0.8}, at={(0.95, 0.3)}},
axis background/.style={fill=gray!12} 
]
\addplot [color=mycolor1, mark=asterisk, mark options={solid, mycolor1}, thick]
  table[row sep=crcr]{%
20  0.09579987\\
40  0.10846132\\
60  0.08502769\\
80  0.13093119\\
100 0.08254737\\
120 0.09207891\\
140 0.12210875\\
160 0.12629881\\
180 0.10889841\\
200 0.07851962\\
};

\addplot [color=mycolor3, mark=o, mark options={solid, mycolor3}, thick]
  table[row sep=crcr]{%
20  0.10266352\\
40  0.11815792\\
60  0.1059098\\
80  0.10846406\\
100 0.05592963\\
120 0.06661392\\
140 0.0915613\\
160 0.01885668\\
180 0.010101937\\
200 0.011656656\\
};
\end{axis}
\end{tikzpicture}%
\end{subfigure}
\begin{subfigure}[b]{0.24\textwidth}
 \centering 
%
%
\definecolor{mycolor1}{rgb}{0.0, 0.44, 1.0}%
\definecolor{mycolor2}{rgb}{0.65000,0.22500,0.09800}%
\definecolor{mycolor3}{rgb}{0.82, 0.1, 0.26}%

\begin{tikzpicture}

\begin{axis}[%
label style={font=\small},
tick label style={font=\small},
y tick label style={/pgf/number format/.cd, fixed,fixed zerofill,precision=2,/tikz/.cd},
width=\figurewidth,
height=\figureheight,
at={(0\figurewidth,0\figureheight)},
scale only axis,
xmin=20,
xmax=200,
xlabel style={font=\color{white!15!black}},
xlabel={Epoch},
ymin=0.,
ymax=0.08,
grid, 
grid style={line width=.15pt, draw=white!0}, 
ylabel style={font=\color{white!15!black}},
axis background/.style={fill=white},
legend style={legend cell align=left, align=left, draw=white!15!black, nodes={scale=0.8}, at={(0.95, 0.3)}},
axis background/.style={fill=gray!12} 
]
\addplot [color=mycolor1, mark=asterisk, mark options={solid, mycolor1}, thick]
  table[row sep=crcr]{%
20  0.04851045\\
40  0.06042966\\
60  0.04950064\\
80  0.07384931\\
100 0.03275994\\
120 0.0087133\\
140 0.05599351\\
160 0.03967444\\
180 0.0642963\\
200 0.04819204\\
};

\addplot [color=mycolor3, mark=o, mark options={solid, mycolor3}, thick]
  table[row sep=crcr]{%
20  0.01239944\\
40  0.01168074\\
60  0.00508785\\
80  0.00266131\\
100 0.01141543\\
120 0.0161207\\
140 0.00294587\\
160 0.01296246\\
180 0.00369718\\
200 0.00350064\\
};
\end{axis}
\end{tikzpicture}%
\end{subfigure}
\begin{subfigure}[b]{0.24\textwidth}
 \centering 
%
%
\definecolor{mycolor1}{rgb}{0.0, 0.44, 1.0}%
\definecolor{mycolor2}{rgb}{0.65000,0.22500,0.09800}%
\definecolor{mycolor3}{rgb}{0.82, 0.1, 0.26}%

\begin{tikzpicture}

\begin{axis}[%
label style={font=\small},
tick label style={font=\small},
y tick label style={/pgf/number format/.cd, fixed,fixed zerofill,precision=2,/tikz/.cd},
width=\figurewidth,
height=\figureheight,
at={(0\figurewidth,0\figureheight)},
scale only axis,
xmin=20,
xmax=200,
xlabel style={font=\color{white!15!black}},
xlabel={Epoch},
ymin=0.,
ymax=0.08,
grid, 
grid style={line width=.15pt, draw=white!0}, 
ylabel style={font=\color{white!15!black}},
axis background/.style={fill=white},
legend style={legend cell align=left, align=left, draw=white!15!black, nodes={scale=0.8}, at={(0.95, 0.3)}},
axis background/.style={fill=gray!12} 
]
\addplot [color=mycolor1, mark=asterisk, mark options={solid, mycolor1}, thick]
  table[row sep=crcr]{%
20  0.00926806\\
40  0.06024201\\
60  0.02983152\\
80  0.03740557\\
100 0.0151438\\
120 0.00857262\\
140 0.0265988\\
160 0.03509023\\
180 0.03130621\\
200 0.06951205\\
};

\addplot [color=mycolor3, mark=o, mark options={solid, mycolor3}, thick]
  table[row sep=crcr]{%
20  0.02655604\\
40  0.00793428\\
60  0.01082472\\
80  0.00216955\\
100 0.01893129\\
120 0.00871699\\
140 0.00572561\\
160 0.00735488\\
180 0.00564814\\
200 0.00234379\\
};
\end{axis}
\end{tikzpicture}%
\end{subfigure}
  \caption{Performance comparison with random search. (Top row) The progression of average performance in top-5 models for different search methods, \textit{i.e.}, AutoOD (red lines with circles), and random search (blue lines with asterisks). (Middle row and bottom row) The mean and standard deviation of model performances in every $20$ epochs along the search progress.} 
  \label{fig:progression}
\end{figure*}

\subsubsection{Comparison Against Traditional NAS}
Instead of applying the policy gradient based search strategy, one can use random search to find the best model. Although this baseline seems simple, it is often hard to surpass~\cite{bergstra2012random,li2019random}. We compare AutoOD with random search to answer the research question \textbf{Q3}. 
The quality of the search strategy can be quantified by the following three metrics: (1) the average performance of the top-$5$ models found so far, (2) the mean performance of the searched models in every $20$ epochs, (3) the standard deviation of the model performance in every $20$ epochs. From Fig.~\ref{fig:progression}, we can observe that: Firstly, our proposed search strategy is more efficient to find the well-performed models during the search process. As shown in the first row of Fig.~\ref{fig:progression}, 
the performance of the top-$5$ models found by AutoOD consistently outperform the random search. The results also show that not only the best model of our search strategy is better than that of random search, but also the improvement of average top models is much more significant. This indicates that AutoOD explores better models faster than the random search. Secondly, 
there is a clear increasing tendency in the mean performance of AutoOD, which can not be observed in random search. It indicates that our search controller can gradually find better strategies from the past search experiences along the learning process, while the random search's controller has a relatively low chance to find a good child model.
Thirdly, compared with the random search, there is a clear dropping of standard deviation 
along the search process. It verifies 
that our search strategy provides a more stable search process.

\begin{figure*}[ht]
  \centering
  \includegraphics[width=\linewidth]{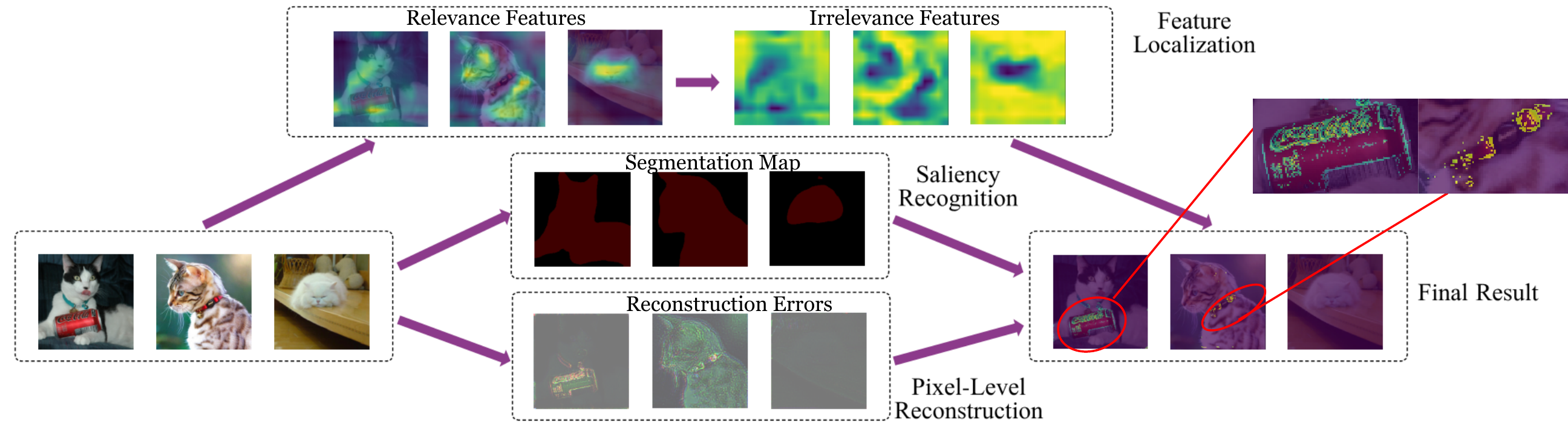}
  \caption{A case study of outlier segmentation. The pixel-level reconstruction is regularized by conducting pixel-level collaboratively with the results of feature localization and saliency recognition. AutoOD accurately identifies the outlier regions (cola bin and collar) of the target object (cat) in the first two images, and treats the third target object (cat) as outlier-region-free.}
  \vspace{5pt}
  \label{fig:cat}
\end{figure*}
\subsection{Case Study}
To answer the research question \textbf{Q4}, we provide further analysis for the pixel-level defect region segmentation task, to get some insights about how to further improve the detection performance in more complicated real-world settings. To make the outlier sections and the rest sections more distinguishable in the latent space, we use the reconstruction error to learn intrinsic representation for positive samples to extract common patterns. Yet, it is hard to directly apply AutoOD into more complicated, real-world settings. Due to the pure data-driven strategy, the reconstruction based denoters might be misled by background noises, other objects, or irrelevance features. We hereby introduce two strategies into AutoOD for regularization without increasing the model complexity.

\noindent \textbf{Saliency refinement via target object recognition.} We introduce a mask map $s$ into the reconstruction based denoters from pre-trained models. It is used for localizing and identifying target salient object, in order to eliminate the negative effect caused by background noises and other objects in the same image. In our experiment, we introduce $s$ from DeepLabV3~\cite{chen2017rethinking}, which is pre-trained on PASCAL VOC 2012~\cite{pascal-voc-2012}. 

\noindent \textbf{Feature augmentation via  gradient-based localization.} In order to amplify the contribution of the irrelevance features from the salient object, we introduce the feature augmentation map to re-weight the reconstruction result. We also introduce a coarse localization map to highlight the irrelevance regions in the image from an interpretability perspective. Feature importance is reflected as gradients signal via backpropagation. Here, we follow the interpretation method from Grad-CAM~\cite{selvaraju2017grad}, which designed to highlight important features, and pre-trained on VGG-16~\cite{simonyan2014very}. The feature augmentation map is defined as opposite to the Grad-CAM: $ \frac{1}{mn} \sum_{m} \sum_{n}  \left(1-\frac{\partial y_{i}}{\partial A_{m n}}\right)$.

After the two steps above, we reweight the reconstructions: 
\begin{equation}
    \alpha_i=\left\|g\left(f\left(x_{i} ; \mathcal{W}\right)\right)-x_{i}\right\|_{2}^{2} \odot \frac{1}{mn} \sum_{m} \sum_{n}  \left(1 -\frac{\partial y_{i}}{\partial A_{m n}}\right) \odot \textrm{s}_{i}
\end{equation}
where $x_{i} \in \mathrm{R}^{m \times n}$ is a training sample and $y_{i}$ is its target object class. $f(\cdot), g(\cdot)$ denotes encoder-decoder structures produced by AutoOD, $A$ denotes the feature map activation of a latent layer.
We use a real-world dataset CAT~\cite{catset} for illustration. We find the optimal model via AutoOD and get the pixel-level reconstruction map. Then, we further refine the map via saliency recognition and feature localization strategies. As can be seen from Fig.~\ref{fig:cat}, AutoOD achieves better visualization results after applying the reweighting tricks discussed above. We also observe that the model can successfully identify the outlier regions (cola bins, collars) within the salient objects (kitties). In the meanwhile, it reduces the effect caused by the background noises and irrelevant objects.

\section{Related Work}
\label{sec:related}
In this section, we review the related work on Neural Architecture Search (NAS). Recently, NAS has attracted increasing research interests. Its goal is to find the optimal neural architecture in a predefined search space to maximize the model performance on a given task. Designing a NAS algorithm requires two key components: the search space and the search strategy (optimization algorithm)~\cite{Elsken2019NeuralAS}. 

The search space defines which architectures can be represented in principles. The existing work of search space follows two trends: the macro and micro search~\cite{pham2018efficient,li2019pyodds,li2020pyodds}. The macro search provides an exhaustive-architecture search space to encourage the controller to explore the space and discover novel architectures, while the micro search inductively limits the search space to accelerate the search process. The choice and the size of the search space determine the difficulty of the optimization problem. Yet, even for the case of the search space based on a single cell, it is still a challenging problem due to the discrete search space and the curse of high-dimensionality (since more complex models tend to perform better, resulting in more design
choices)~\cite{Elsken2019NeuralAS}. Thereby, incorporating prior knowledge about the typical properties of architectures well-suited for a task can significantly reduce the size of the search space and simplify the search process. ~\cite{li2019random} has validated the importance of the search space in the search process. With extensive experimental reproducibility studies, a task-tailored, carefully-designed search space plays a more important role than the other search strategies. Recent works have proposed tailored search spaces with their applications, including image segmentation~\cite{liu2019auto}, adversarial training~\cite{gong2019autogan}, and augmentation strategies~\cite{cubuk2019autoaugment}. To the best of our knowledge, our proposed AutoOD describes the first attempt to design the search space specifically customized to the outlier detection task. AutoOD uses the micro search space to keep consistent with previous works. Yet the contribution of AutoOD in search space is to design a hierarchical, general-purpose search space, including global settings for the whole model, and local settings in each layer independently. Moreover, our proposed search space not only covers the hyperparameters as architecture configurations, such as the size of convolutional kernels and filters in each layer, but also incorporates the definition-hypothesis and its corresponding objective function.  

The search strategy focuses on how to explore the search space. Recent approaches include reinforcement learning (RL)~\cite{pham2018efficient,zoph2016neural,cai2018proxylessnas}, 
Bayesian optimization~\cite{jin2019auto}, and gradient-based methods~\cite{liu2018darts,brock2017smash,liu2018progressive}. 
Although these methods have improved upon human-designed architectures, directly borrowing existing NAS ideas from image classification to outlier detection will not work. Due to the imbalanced data, the search process becomes more unstable in outlier detection tasks~\cite{swirszcz2016local}. As an internal mechanism in the traditional NAS, weight sharing, also introduces the inductive bias in the search process which intensifies the tendency~\cite{chu2019fairnas}. 
Weight sharing~\cite{pham2018efficient} is proposed to transfer the well-trained weight before to a sampled architecture, to avoid training the offspring architecture from scratch. Recent research has validated that the weight sharing mechanism makes the architectures who have better initial performance with similar structures more likely to be sampled~\cite{chu2019fairnas}, which leads to misjudgments of the child model's performance. Our work builds upon RL-based method, which uses a recurrent neural network controller to choose blocks from its search space. Beyond that, we propose a curiosity-guided search strategy to stabilize the search process via encouraging the controller to seek out unexplored regions in the search space. Our search strategy formulates the search process as a classical exploration-exploitation trade-off. On one hand, it is encouraged to find the optimal child model more efficiently; on the other hand, it avoids the premature convergence to a sub-optimal region due to the inductive bias or insufficient search.

\section{Conclusions}
\label{sec:con}
In this paper, we investigated a novel and challenging problem of automated deep model search for outlier detection. Different from the existing Neural Architecture Search methods that focus on discovering effective deep architectures for supervised learning tasks, we proposed \modele, an automated unsupervised outlier detection framework, which aims to find an optimal neural network model within a predefined search space for a given dataset. \model builds on the theory of curiosity-driven exploration and self-imitation learning. It overcomes the curse of local optimality, the unfair bias, and inefficient sample exploitation problems in the traditional search methods. We evaluated the proposed framework using extensive experiments on eight benchmark datasets for instance-level outlier sample detection and pixel-level defect region segmentation. The experimental results demonstrated the effectiveness of our approach. 
\bibliographystyle{ieeetr}
\bibliography{sample-base}


\end{document}